\documentclass[10pt,twocolumn,letterpaper]{article}

\pdfoutput=1
\usepackage[pagenumbers]{cvpr}      % To produce the REVIEW version
\usepackage{url}            % simple URL typesetting
\usepackage{booktabs}       % professional-quality tables
\usepackage{amsfonts}       % blackboard math symbols
\usepackage{nicefrac}       % compact symbols for 1/2, etc.
\usepackage{microtype}      % microtypography
\usepackage{xcolor}         % colors
\usepackage{graphicx}
\usepackage{amsmath}
\usepackage{amssymb}
\usepackage{booktabs}
\usepackage{array}
\usepackage{colortbl}
\usepackage{multirow} % For multirow feature
\usepackage{arydshln} % For dashed lines
\usepackage{hhline} % For double horizontal lines
\usepackage{amssymb} % For checkmark symbol
\usepackage{wrapfig}
\usepackage{graphicx}
\usepackage{amsmath}
\usepackage{subcaption}
\usepackage[export]{adjustbox}
\usepackage{lipsum}
\usepackage{moresize}
\usepackage[accsupp]{axessibility}
\usepackage{indentfirst}

\definecolor{cvprblue}{rgb}{0.21,0.49,0.74}
\usepackage[pagebackref,breaklinks,colorlinks,allcolors=cvprblue]{hyperref}

\newcolumntype{L}{@{}>{\kern\tabcolsep}l<{\kern\tabcolsep}}
\newcolumntype{H}{>{\setbox0=\hbox\bgroup}c<{\egroup}@{}}

\makeatletter
\newcommand*\bigcdot{\mathpalette\bigcdot@{1.2}}
\newcommand*\bigcdot@[2]{\mathbin{\vcenter{\hbox{\scalebox{#2}{$\m@th#1\bullet$}}}}}
\makeatother

\DeclareMathOperator*{\argmax}{arg\,max}

\definecolor{LAA-Net}{HTML}{FFC300}
\definecolor{SBI}{HTML}{839788}
\definecolor{CADDM}{HTML}{CBC0D3}
\definecolor{ViT}{HTML}{CD8987}
\definecolor{Swin}{HTML}{184E77}

\newtheorem{Definition}{Definition}

\usepackage[capitalize]{cleveref}
\crefname{section}{Sec.}{Secs.}
\Crefname{section}{Section}{Sections}
\Crefname{table}{Table}{Tables}
\crefname{table}{Tab.}{Tabs.}

\def\paperID{16999} % *** Enter the Paper ID here
\def\confName{CVPR}
\def\confYear{2025}

\title{FakeFormer: Efficient Vulnerability-Driven Transformers for Generalisable Deepfake Detection}

\author{Dat NGUYEN$^{\triangleleft}$, Marcella ASTRID$^{\triangleleft}$, Enjie GHORBEL$^{\triangleleft,\rtimes}$, Djamila AOUADA$^{\triangleleft}$ \\
CVI$^2$, SnT, University of Luxembourg$^{\triangleleft}$ \\
Cristal Laboratory, National School of Computer Sciences, University of Manouba$^{\rtimes}$ \\
{\tt\small \{dat.nguyen, marcella.astrid, enjie.ghorbel, djamila.aouada\}@uni.lu}
}

\begin{document}
\maketitle

\begin{abstract}
Recently, Vision Transformers (ViTs) have achieved unprecedented effectiveness in the general domain of image classification. Nonetheless, these models remain underexplored in the field of deepfake detection, given their lower performance as compared to Convolution Neural Networks (CNNs) in that specific context. In this paper, we start by investigating why plain ViT architectures exhibit a suboptimal performance when dealing with the detection of facial forgeries. Our analysis reveals that, as compared to CNNs, ViT struggles to model localized forgery artifacts that typically characterize deepfakes. Based on this observation, we propose a deepfake detection framework called FakeFormer, which extends ViTs to enforce the extraction of subtle inconsistency-prone information. For that purpose, an explicit attention learning guided by artifact-vulnerable patches and tailored to ViTs is introduced. Extensive experiments are conducted on diverse well-known datasets, including FF++, Celeb-DF, WildDeepfake, DFD, DFDCP, and DFDC. The results show that FakeFormer outperforms the state-of-the-art in terms of generalization and computational cost, without the need for large-scale training datasets. 
% \texttt{The code is provided in supplementary materials and will be publicly released}.
The code is available at \url{https://github.com/10Ring/FakeFormer}.
\end{abstract}

\section{Introduction}
\label{sec:intro}

\begin{figure}
    \centering
    \includegraphics[width=0.9\linewidth]{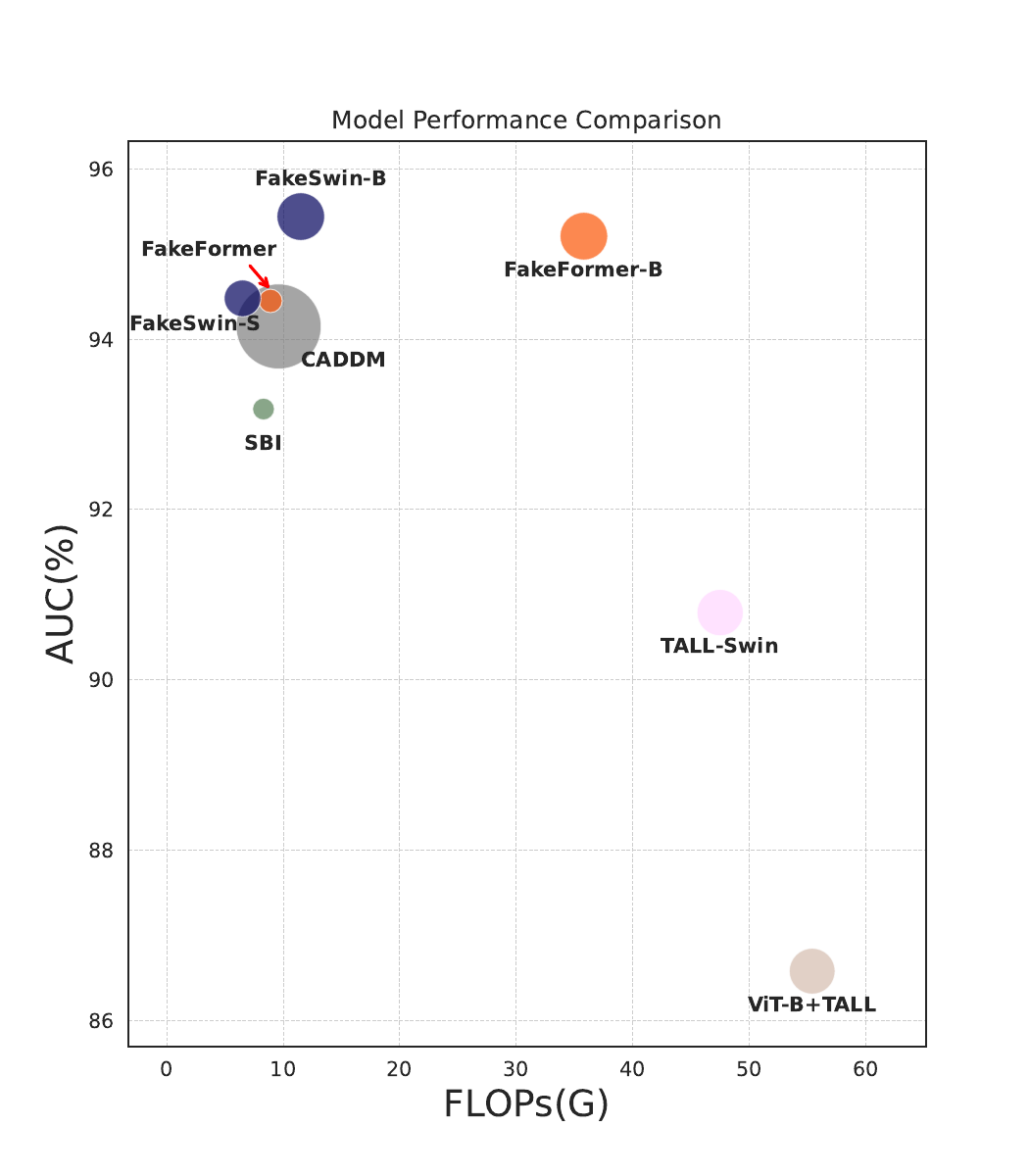}
    \vspace{-2mm}
    \caption{Comparison of FakeFormer (and FakeSwin) to existing methods including SBI~\cite{sbi}, CADDM~\cite{caddm}, and Transformer-based approaches~\cite{tall_swin}, namely TALL-Swin and ViT-B+TALL, in terms of model size, FLOPs, and AUC. The size of each bubble represents the number of model parameters. All methods are trained on FF++~\cite{ff++} and tested on CDF2~\cite{celeb_df}.}
    \label{fig:motivation}
    \vspace{-3mm}
\end{figure}

In the era of deep learning, it has never been easier to generate ultra-realistic facial images, undetectable to the naked eye. Such forged visual data, commonly called deepfakes~\cite{deepfake1, deepfake2}, are produced by manipulating the identity or the expression of a subject using generative deep neural networks such as Generative Adversarial Networks (GAN)~\cite{gan}. Although beneficial in several applications, deepfakes can also be harmful to society if employed for malicious purposes such as spreading misinformation and fraud. 

%uses Deep Neural Networks (DNNs) to manipulate facial identities or expressions that can not be simply distinguished by human naked eyes. 
% Such fascinating techniques are beneficial for several applications such as entertainment, movie industries, etc; 
%Besides many beneficial incentives witnessed, it is uncontrollably abused for malicious purposes. This phenomenon raises numerous critical societal concerns, e.g. spreading political vulgarization, committing fraud transactions, defaming celebrities, etc.
To face this growing threat, the research community is making ceaseless efforts to introduce approaches for automatically detecting deepfakes. Currently, the most successful deepfake detection methods are mostly relying on Convolution Neural Networks (CNNs)~\cite{caddm, lipforensics, qual_agnostic, altfreezing, sbi, cstency_learning, sladd,  laa_net, localRL, ost, ucf, aunet, sadd, fgi}. Paradoxically, thanks to their simplicity and scalability, Vision Transformers~\cite{ViT, deit} (ViTs) are becoming increasingly popular in other visual classification tasks, often outperforming CNN architectures. However, ViT architectures remain relatively underexplored in the field of deepfake detection, given their lower performance as compared to CNNs in that context~\cite{ict, m2tr}. 

In this paper, we aim to: \textit{(1) investigate the reasons behind this drop in performance with respect to CNNs} and \textit{(2) suggest accordingly adequate yet simple strategies for enhancing the performance of ViT architectures in the context of deepfake detection, while benefiting from their efficiency}. 
Specifically, we first experimentally analyze the capability of ViTs to capture localized artifacts typically characterizing deepfakes, as compared to CNNs. Our analysis suggests that ViTs struggle more in this regard because they tend to focus on global context~\cite{ViT, twin, nat, swin, PVT}. % Although reducing the patch size of the ViTs' input mitigates this problem, it quadratically increases computational complexity due to the self-attention mechanism.

% , while deepfake forgeries become less obvious to detect~\cite{celeb_df, dfdc, fsganv1, fsganv2}; and 2) are known to be data-hungry~\cite{ViT, vitpose, ViTneedlargescale1, deit}, while the available deepfake dataset is limited.}
% Specifically, we first experimentally investigate why ViT-based approaches struggle to model localized artifacts in comparison to CNN-based methods, which might make them less suitable for detecting manipulated facial images, in particular, high-quality deepfakes. \textcolor{red}{We believe that this outcome happens for two reasons: 1) generative methods are advancing, making forgeries less obvious to detect~\cite{celeb_df, dfdc, fsganv1, fsganv2}, and 2) the limited availability of large-scale datasets in the field of deepfake detection further restricts the ability to fully utilize the capacity of ViTs~\cite{ViT, vitpose, ViTneedlargescale1, deit}.}
Therefore, to address this issue, 
% Second, 
% to address this issue without relying on costly annotation efforts \textcolor{blue}{and without significantly increasing the computational complexity}, 
we propose to inject a local attention mechanism in ViTs by explicitly detecting vulnerable patches using blending-based data synthesis techniques such as~\cite{sbi,fxray}. For that purpose, we extend the notion of vulnerable points~\cite{laa_net}, which represent the pixels that are the most impacted by the blending operation, to vulnerable patches. These patches are the group of pixels that are the most likely to incorporate blending artifacts and can be used in conjunction with any transformer-like architecture. This attention module is referred to as Learning-based Local Attention mechanism (L2-Att). 
Even though the idea of attention to vulnerabilities has been explored in~\cite{laa_net}, our experimental study shows that generalizing this approach to transformers contributes to:

\begin{enumerate}
    \item \textbf{Enhanced generalization performance:} As reflected in Figure~\ref{fig:motivation}, higher generalization capabilities are achieved by different transformer-based architectures enclosing the proposed L2-Att strategy, namely FakeFormer and FakeSwin. Specifically, the FakeSwin is a variant of FakeFormer where Swin~\cite{swin} is used as a backbone. The notations -S and -B refer to the backbone size (small and base, respectively)\footnote{ More details are reported in Section~\ref{sec:experiment}.}.
    \item \textbf{Friendly computational cost and reduced model size}: As illustrated in Figure~\ref{fig:motivation}, although achieving impressive performance, the proposed transformer models cost a relatively low number of FLOPs, and necessitate fewer parameters as compared to other methods.
    % \item  \textbf{Reduced model size}: Although achieving impressive performance, FakeFormer necessitates a lower number of parameters with respect to other methods as shown in Figure~\ref{fig:motivation}.
    \item \textbf{Mitigating the need for large-scale datasets}: ViTs are known to be data-hungry~\cite{ViT, vitpose, ViTneedlargescale1, deit}. Therefore, by employing generic blending-based synthesis methods as~\cite{fxray, sbi} to define pseudo-vulnerable patches, we mitigate this problem.
    % the usual ViT's need for large-scale training.
\end{enumerate}
%better generalization performance but also helps reduce the number of parameters and FLOPs, as shown in Figure.~\ref{fig:motivation}. Moreover, we further demonstrate that by employing generic blending-based synthesis methods as~\cite{fxray, sbi} to define pseudo-vulnerable patches, we mitigate the usual ViT's need for large-scale training.
This is confirmed by extensive experiments conducted on seven challenging datasets including FF++~\cite{ff++}, two versions of CDF~\cite{celeb_df}, DFD~\cite{dfd}, DFW~\cite{wdf}, DFDCP~\cite{dfdcp}, and DFDC~\cite{dfdc}. Our method outperforms state-of-the-art approaches, including both CNN-based and ViT-based methods~\cite{ict, dfdt, m2tr}, although the latter \cite{ict, dfdt, m2tr} usually rely on a significantly larger training set.
%Additionally, we make attempts to simulate the real-life conditions by combining all these datasets into one, namely the ``Combined'' dataset. 
%The reported results demonstrate that FakeFormer achieves superior performance as compared to the state-of-the-art (SOTA) in terms of both speed and performance.

\vspace{1mm}
\noindent\textbf{Paper Organization.} Section~\ref{sec:related_w} discusses related work. Section~\ref{sec:investigation} analyzes the performance gap between ViTs and CNNs in deepfake detection. The proposed approach called FakeFormer is presented in Section~\ref{sec:method}. Section~\ref{sec:experiment} and Section~\ref{sec:limitation} discuss the experiments and the limitations, respectively. Section~\ref{sec:conclu} concludes this work and suggests future directions.

\begin{figure}
    \centering
    \includegraphics[width=0.75\linewidth]{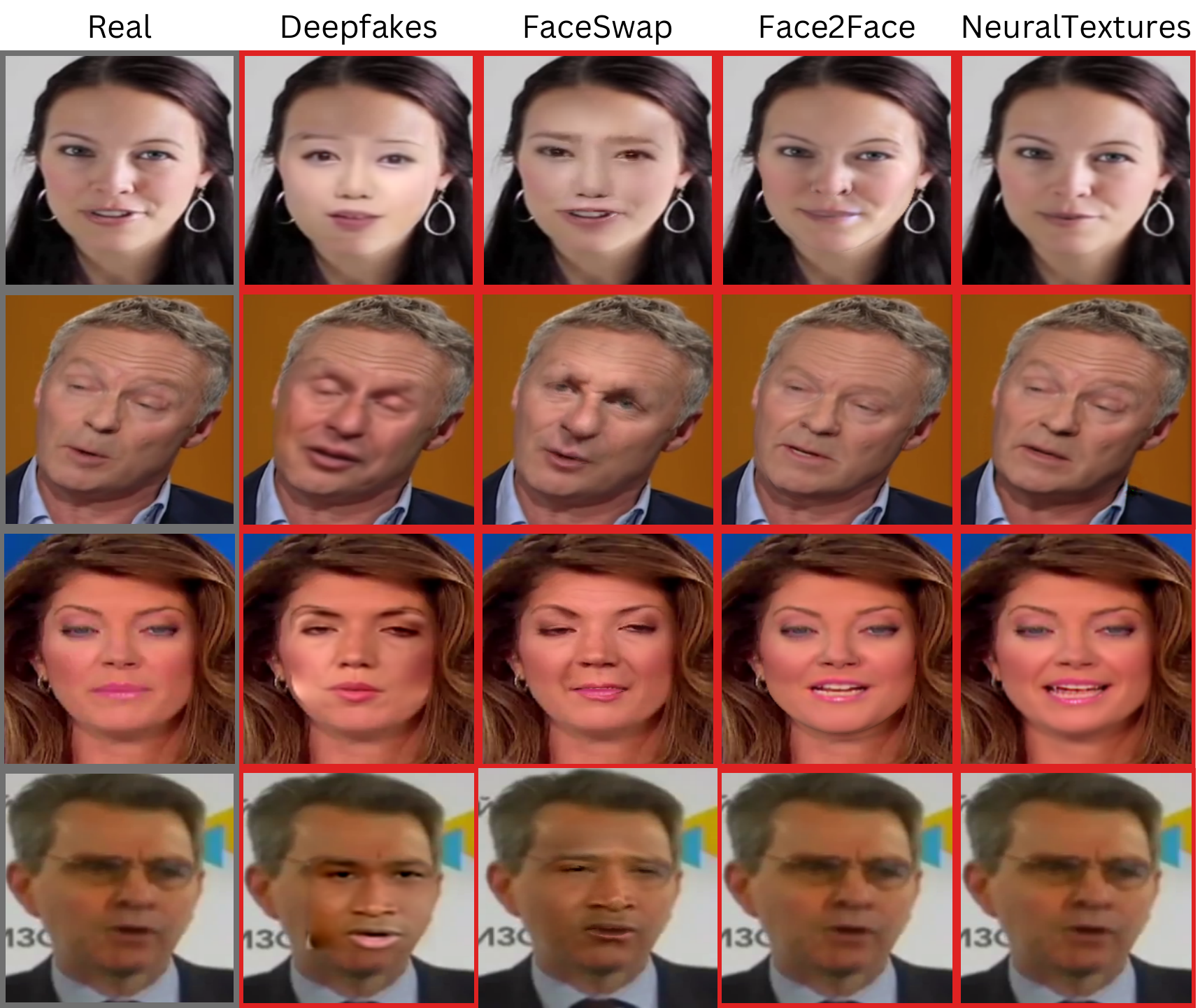}
    \caption{Examples are randomly selected to illustrate the four types of deepfakes in common FF++~\cite{ff++} dataset. It can be observed that Face2Face~\cite{face2face} and NeuralTextures~\cite{neutex} exhibit more subtle artifacts.}
    \label{fig:locality_levels}
    \vspace{-2mm}
\end{figure}

\begin{figure*}
    \begin{subfigure}[b]{0.32\linewidth}
        \centering
        \includegraphics[width=\linewidth, valign=t]{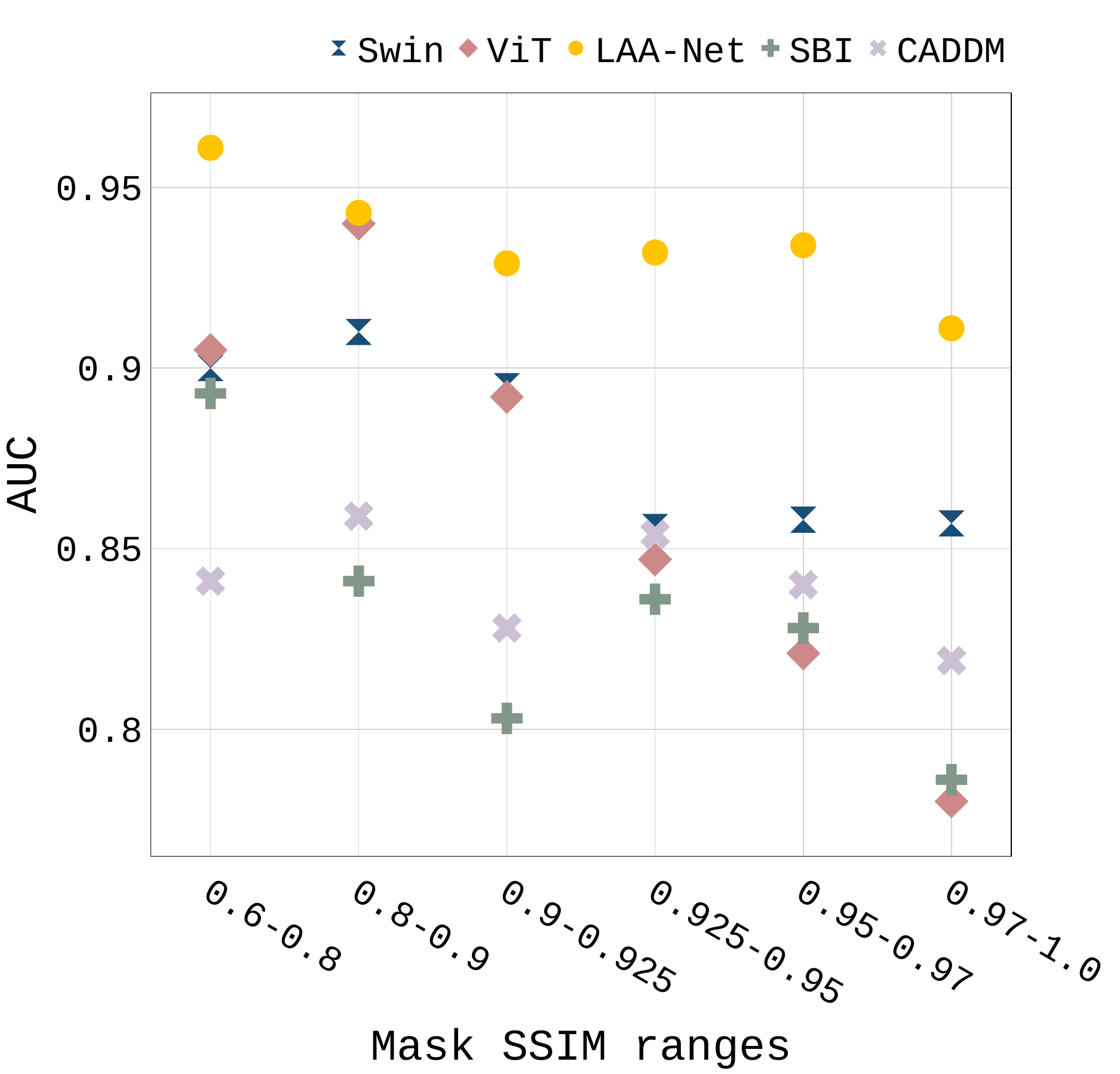}
        \caption{}
        \label{fig:ssim_ranges}
    \end{subfigure}
    \hfill
    \begin{subfigure}[b]{0.67\linewidth}
        \centering
        \includegraphics[width=\linewidth, valign=t]{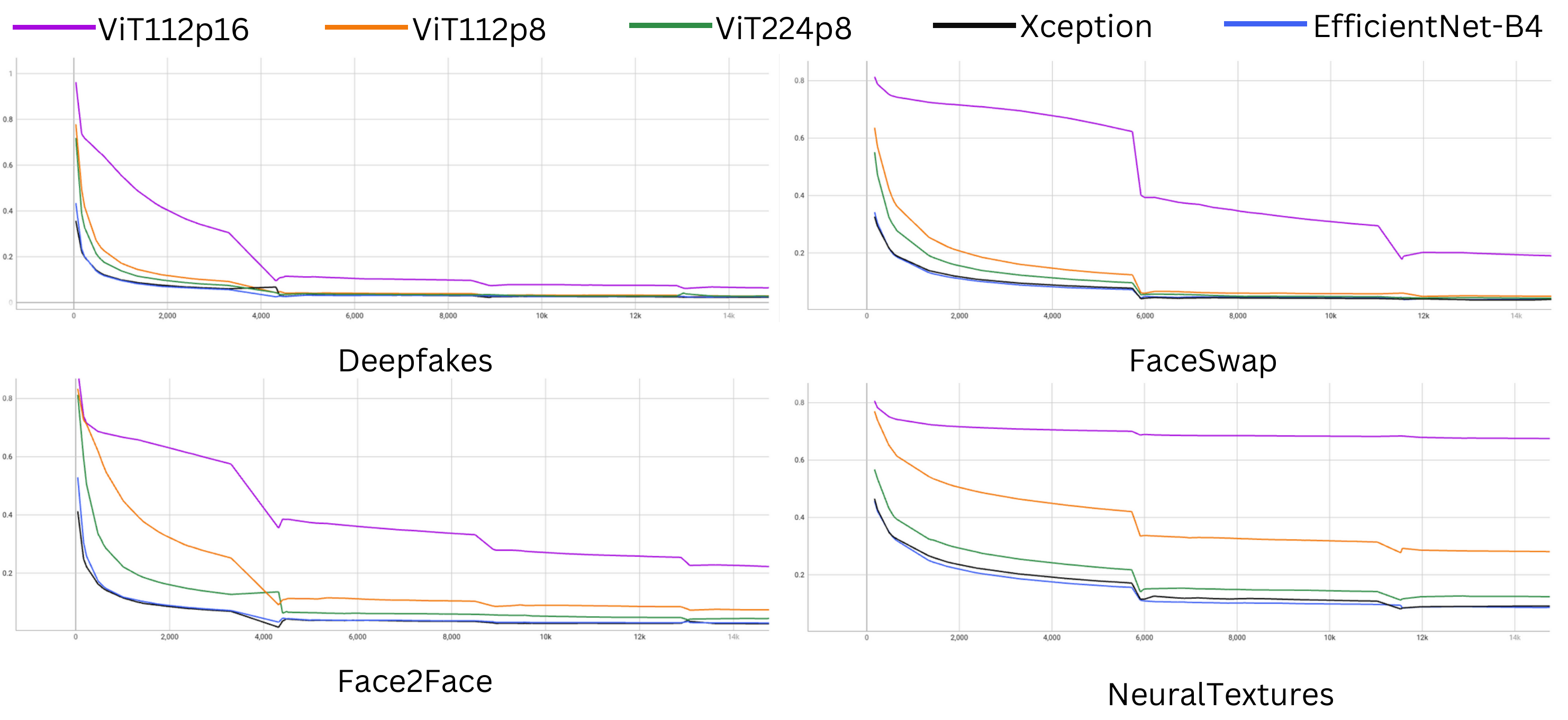}
        \caption{}
        \label{fig:ViT_CNN_losses}
    \end{subfigure}%
    \vspace{-3mm}
    \caption{Experiments to analyze the capability of transformer-based networks in deepfake detection: (a) Performance comparison of transformer-based architectures (ViT\cite{ViT}\textcolor{ViT}{$\bigcdot$}, Swin\cite{swin}\textcolor{Swin}{$\bigcdot$}) and three CNN-based methods (LAA-Net\cite{laa_net}\textcolor{LAA-Net}{$\bigcdot$}, CADDM\cite{caddm}\textcolor{CADDM}{$\bigcdot$}, SBI \cite{sbi}\textcolor{SBI}{$\bigcdot$}) across different ranges of Mask-SSIM~\cite{mssim_pose}. All methods are trained on FF++~\cite{ff++} and tested on CDF2~\cite{celeb_df}. (b) Evolution of the training loss using ViT under different configurations (variation of input resolution and patch size), XceptionNet~\cite{xception} and EfficientNet-B4~\cite{efn_net}, across four types of deepfakes in FF++~\cite{ff++}.}
    \vspace{-4mm}    
\end{figure*}

\section{Related Work}
\label{sec:related_w}

\paragraph{Deepfake Detection with CNNs.}

%\textcolor{red}{Paragraph 1: CNNs and their weakness} 
%\begin{itemize}
 %   \item \textcolor{red}{Discuss and cite several learning-based CNN models (collaborative learning, disentanglement learning, online-test time training, adversarial learning, etc) => heavy + slow.}
  %  \item \textcolor{red}{Mentioning LAA-Net + concept of vulnerable points => limitations of LAA-Net (bigger no. parameters, higher FLOPs, multi-branches + the Decoder cause high latency) => vulnerable patches compatible with ViTs embedding outputs, LGAM is a very lightweight overhead computation module.}
%\end{itemize}

In the field of deepfake detection,  Convolution Neural Networks (CNNs) are often favored over other types of architectures. Pioneer works in deepfake detection have predominantly employed established CNN architectures such as XceptionNet~\cite{xception} as binary classifiers~\cite{ff++, mesonet, bin_dfs}. However, these methods struggle to generalize to unseen manipulation methods. To address this challenge, a wide range of strategies have been investigated~\cite{caddm, qual_agnostic,ete_recons, ucf, sbi, fxray, untag} such as disentanglement learning~\cite{ete_recons, ucf}, multi-task learning~\cite{sladd, laa_net, cstency_learning}, and pseudo-fake synthesis~\cite{sbi, fxray, cstency_learning, ost, untag}. With the advances in generative modeling, deepfakes are constantly gaining in realism, resulting in invisible localized artifacts. The aforementioned methods are exposed to the risk of becoming obsolete. Indeed, they still rely on standard CNNs, consequently facing a loss of local information through successive convolution layers~\cite{multi-attentional,sfdg}. For capturing more effectively low-level features, methods with implicit attention strategies have been proposed~\cite{multi-attentional,sfdg,rfm}, but have shown poor generalization~\cite{laa_net}. Recently, Nguyen \etal ~\cite{laa_net} argued that to ensure robustness to high-quality deepfakes while maintaining good generalization capabilities, an explicit attention mechanism within a multi-task learning framework is needed.  In particular, they introduce LAA-Net, a network guided to focus on vulnerable points (defined as the pixels that are the most likely to incorporate blending artifacts). To simulate vulnerable pixels, blending-based data synthesis techniques are leveraged. However, although LAA-Net achieves state-of-the-art performances, it is sensitive to noise and is based on a cumbersome CNN that is computationally costly, thereby limiting its practical deployment.
In this paper, we investigate how to adapt such an explicit attention mechanism to ViT-based architectures to improve the effectiveness of transformer-based deepfake detection methods and benefit from more efficient models.

\vspace{1mm}
\noindent\textbf{Deepfake Detection with ViTs.}
%\textcolor{red}{The success of ViTs in many vision tasks, expecting it to become the alternative to CNNs} 
%\begin{itemize}
 %   \item \textcolor{red}{The desire to discover the prosperity of ViTs in generic DF detection tasks. FTCN + LTTD + ... for video, heavy for practical use, deployment. TALL introduces a novel thumbnail for video DF detection in a spatial manner, requiring a high-resolution input => utilize Swin, a hierarchical SOTA transformer-based backbone.}
  %  \item \textcolor{red}{Detecting DFs from the perspective of inner-outer face ID consistency, ICT~\cite{ict} proposed an efficient ViT for image-level DF detection. Although achieving desirable inference speed, ICT still requires large-scale datasets for training and witnesses limited generalization performance.}
   % \item \textcolor{red}{M2TR is a multi-branch network that incorporates multiple patch sizes + require very large-scale training => slow + heavy + inpractical.}
    %\item \textcolor{red}{In this paper, introduce FakeFormer, a simple, lightweight, generalizable DF detector for practical use. In addition, explain why ViT less appealing in DF detection + LGAM can be integrated to improve performance in any ViT-based models.}
%\end{itemize}
Vision Transformers (ViTs) have attracted the great interest of the research community, as they are claimed to outperform CNNs in terms of both effectiveness and efficiency. While such a statement holds for several applications~\cite{efdet, ViT, vitpose, mae, deit, detr}, it does not apply to the field of deepfake detection. So far, only a handful of transformer-based deepfake detection methods have been explored in the literature. For instance, ICT~\cite{ict} utilizes a basic ViT to detect deepfakes by assessing identity consistency between inner and outer faces in suspect images. To capture multi-scale features, M2TR~\cite{m2tr} and DFDT~\cite{dfdt} simultaneously process inputs through multiple transformers with different patch sizes. However, despite training on large-scale datasets~\cite{ms_celeb_1m} and/or requiring costly computational resources, these methods still exhibit limited generalization performance compared to CNN-based approaches. This motivates us to investigate in Section~\ref{sec:investigation} the underlying reasons explaining the unsuitability of ViTs for this use-case.

\begin{figure*}
    \centering
    \includegraphics[width=0.9\linewidth]{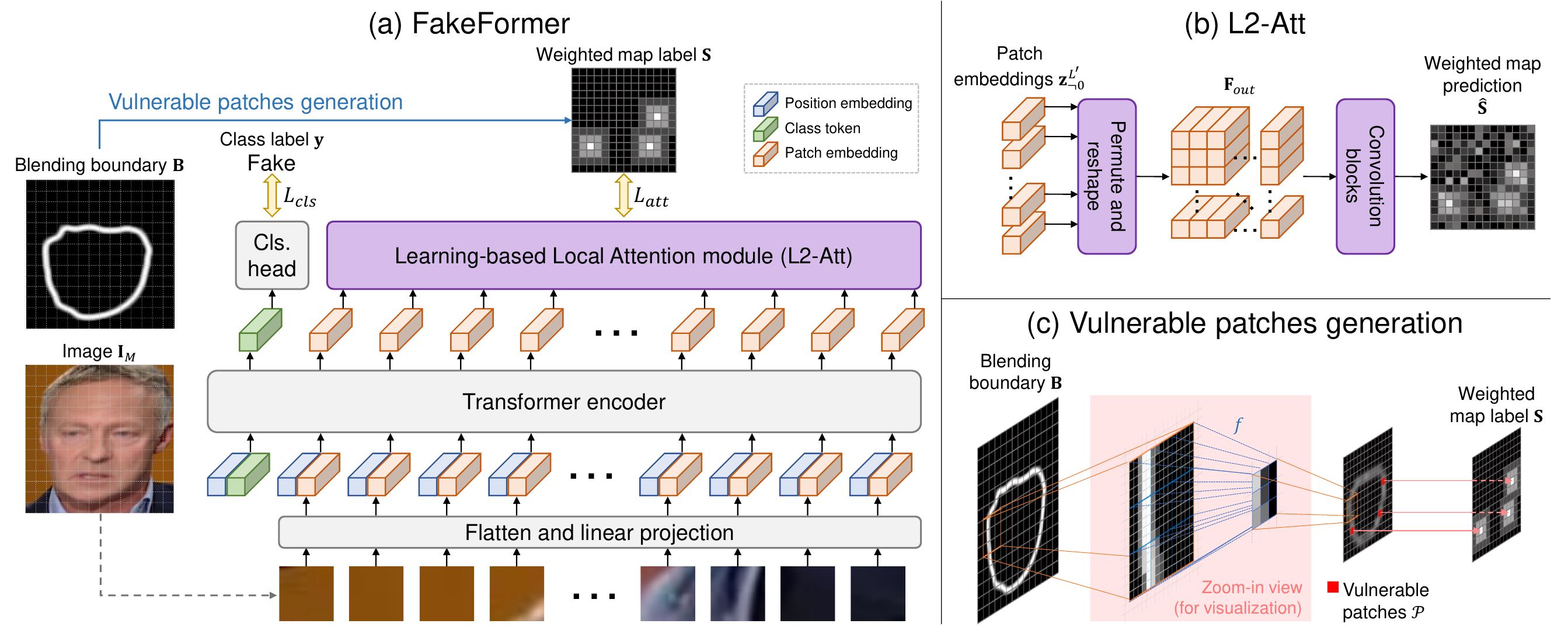}
    \vspace{-2mm}
    \caption{The proposed method: (a) the overall FakeFormer framework, (b) the L2-Att module, and (c) the generation of vulnerable patches.}
    \label{fig:fakeformer_ovv}
    \vspace{-4mm}
\end{figure*}

\section{Vision Transformer and Deepfake Detection: Where is the Gap?}
\label{sec:investigation}

In this section, we investigate the specific challenges associated with the use of plain ViTs in deepfake detection. Our primary hypothesis is as follows: Compared to CNNs, ViTs focus on global image representations~\cite{ViT, twin, nat, swin, PVT}, given their patch-based architecture. Consequently, they struggle to effectively capture local features~\cite{twin, nat} that are crucial for identifying subtle artifacts in deepfakes. 
% Another possible reason is that ViTs generally call for large-scale training datasets~\cite{ViT, deit, ViTneedlargescale1, vitpose}, which are scarce in the deepfake detection domain. These are relatively small compared to those in other fields, such as~\cite{imagenet21k, JFT-300M}. \textcolor{red}{This hypothesis is partly supported by previous works~\cite{m2tr, ict, efn_vit, dfdt} demonstrating the advantages of larger size of data}. 
As a result, CNN-based architectures often outperform ViT-based ones in this context. 
%\textcolor{blue}{One possible reason is the patch size used in ViTs' input can be too large, making it more difficult to capture subtle artifacts. Reducing the patch size can improve the performance of ViTs in detecting deepfakes, despite the increased complexity quadratically}}.
The plausibility of this assumption is investigated by conducting the experiments described below.

% First, as seen in Figure~\ref{fig:ViT_CNN_losses}(a), we compare the performance of CNN-based (LAA-Net~\cite{laa_net}, CADDM~\cite{caddm}, and SBI~\cite{sbi}) and transformer-based (ViT~\cite{ViT} and SWIN~\cite{swin}) methods in different Mask SSIM~\cite{mssim_pose} ranges. Higher Mask SSIM values indicate higher quality of deepfakes, which can also indicate the artifacts are more local. As seen, the performance of ViT drop more significantly with higher Mask SSIM in comparison to several CNN-based methods (LAA-Net and CADDM). SWIN's performance does not drop as much as it can extract more local features with its local window design; however it is still inferior to LAA-Net. These observations can validate hypothesis \#1. Note that, for a fair comparison with SOTA, we train ViT and Swin using the same data synthesis~\cite{sbi} as SBI~\cite{sbi} and LAA-Net~\cite{laa_net}.
\vspace{2mm}
\noindent\textbf{Performance with respect to the quality of deepfakes.} Here, our goal is to quantify the performance of ViTs and CNNs with respect to the quality of the encountered deepfakes. To that aim, we compare in Figure~\ref{fig:ssim_ranges} the performance of CNN-based methods
% ~\footnote{The FLOPs and the number of parameters are presented in Section~\ref{sec:exp}.}
(LAA-Net~\cite{laa_net}, CADDM~\cite{caddm}, and SBI~\cite{sbi}) to transformer-based approaches (Plain ViT~\cite{ViT} and Swin~\cite{swin})\footnote{Note that the small(-S) architecture configuration is employed for both ViT and Swin. The FLOPs and the number of parameters are reported in Section~\ref{sec:experiment}.} across different Mask Self-Similarity Index Measure (SSIM)~\cite{mssim_pose} ranges on the Celeb-DF~\cite{celeb_df} dataset. The Mask SSIM has been commonly used for assessing the quality of deepfakes~\cite{celeb_df, m2tr, laa_net} and is obtained by calculating the SSIM~\cite{ssim_1} score in the facial region between the fake and the corresponding original images. Higher Mask SSIM values indicate a higher perceptual quality of deepfakes~\cite{celeb_df, m2tr, mask-ssim-ref, laa_net}, generally implying the presence of more localized artifacts. It can be noted from Figure~\ref{fig:ssim_ranges} that the performance of ViT drops more significantly for higher Mask SSIM values as compared to CNN-based methods. The performance of Swin, on the other hand, does not deteriorate as much due to its ability to extract low-level features with its local window design; however, it still shows less stability than LAA-Net. These observations support our hypothesis. We specify that for a fair comparison with the state-of-the-art (SOTA), we train ViT and Swin using the same data synthesis~\cite{sbi} used in SBI and LAA-Net. More experiments will be provided in supplementary materials.

\vspace{2mm}
\noindent\textbf{Performance of ViTs with respect to the patch size and the type of deepfakes.}  We investigate whether there exists a correlation between the patch size and the performance of ViTs in deepfake detection. Specifically, we anticipate that smaller patch sizes would help model more localized artifacts. For that purpose, we train a plain ViT with several configurations, by varying both the patch size and the input resolution. Figure~\ref{fig:ViT_CNN_losses} depicts the evolution of the training loss through epochs. The notation ViT$X$p$Y$ in Figure~\ref{fig:ViT_CNN_losses} denotes an input resolution of $X$ with a patch size of $Y$. 
We also compare ViTs to CNNs by training the two most widely-adopted CNNs in deepfake detection~\cite{sbi, sladd, aunet, multi-attentional, sfdg, ff++, ete_recons}, XCeptionNet~\cite{xception} and EfficientNet~\cite{efn_net}. Both ViTs and CNNs are trained on four types of deepfakes from FF++~\cite{ff++}: Deepfakes (DF)~\cite{deepfake}, FaceSwap (FS)~\cite{faceswap}, Face2Face (F2F)~\cite{face2face}, and NeuralTextures (NT)~\cite{neutex} as shown in Figure~\ref{fig:locality_levels}. For the training setups, we train all models for $50$ epochs, using $128$ uniformly extracted from each real/fake video for training and $32$ frames for validation. 
% \textcolor{red}{This significantly reduces the amount of training data compared to previous ViT-based works~\cite{ict, dfdt, m2tr}. Our primary objective is to evaluate the performance of the ViT in deepfake detection under conditions of limited data availability}. 
More details e.g., optimizer, learning-rate scheduler, etc., are provided in the supplementary materials.

As F2F and NT correspond to face reenactment manipulations while FS and DF represent face-swap approaches, it is more likely to observe more subtle artifacts in F2F and NTs. When comparing ViT112p16 and ViT112p8, it can be seen that a ViT with smaller patches exhibits faster convergence compared to those with larger ones. Moreover, increasing the input resolution while conserving the same patch size (see ViT112p8 and ViT224p8) contributes to the amplification of the local information encoded in each patch, also resulting in faster convergence. In both cases, this convergence gap is even more pronounced for more subtle deepfake types such as F2F and NT, indicating the importance of locality in detecting deepfakes with subtle inconsistencies. However, reducing the patch size leads to a quadratically increasing complexity (i.e. the FLOPs are $2.2$G, $8.4$G, and $33.6$G for ViT112p16, ViT112p8, and ViT224p8, respectively). Additionally, it is to note that CNNs converge more rapidly compared to ViTs under different setups (i.e., even with the smallest patch size). This also highlights the fact that CNNs can extract local features more effectively. 
% \textcolor{blue}{Another possible reason why ViTs converge more slowly than CNNs can be the small training dataset (i.e., only a subset of FF++ without augmentation), as ViTs are widely-known to be data-hungry~\cite{ViT, deit, ViTneedlargescale1, vitpose}.}

Hence, we posit that by proposing a mechanism that allows focusing on subtle artifact-prone regions, we can enhance the performance of ViTs for the task of deepfake detection. While some attempts have been made to introduce local ViTs such as Swin~\cite{swin}, we argue that this remains insufficient for effectively detecting deepfakes. As demonstrated for CNNs~\cite{laa_net}, incorporating local features does not guarantee that artifact-prone are considered, highlighting the need to introduce attention strategies for explicitly focusing on localized artifacts.

\section{FakeFormer}
\label{sec:method}
\vspace{-2mm}

Based on the observations made in Section~\ref{sec:investigation}, we propose FakeFormer, a Transformer-based approach integrating an attention mechanism that induces the modeling of subtle inconsistencies. Aside from its effectiveness, FakeFormer is not data-hungry and is computationally efficient. An overview of the proposed approach can be seen in Figure~\ref{fig:fakeformer_ovv}, consisting of a Vision Transformer (ViT) coupled with a Learning-based Local Attention module (L2-Att). L2-Att predicts the locations of vulnerable patches which generalize the notion of vulnerable points introduced in~\cite{laa_net}. FakeFormer is trained using only normal data and utilizes blending-based data synthesis techniques~\cite{fxray,sbi}. It can be noted that we utilize ViT to formulate our methodology for the sake of simplicity. However, our method is compatible with other transformer-based architectures, as demonstrated in Section~\ref{sec:experiment}. In what follows, we depict the different components of FakeFormer.

\subsection{Vision Transformer (ViT)}
\label{subsec:overview}

% \textbf{\textcolor{red}{Problem formulation for FakeFormer:}}

Given an image $\mathbf X \in \mathbb{R}^{C \times H \times W}$ as input, we first reshape it into a sequence of non-overlapping flattened 2D patches, denoted as $\{\mathbf x_i \in \mathbb R^{C.P^2} \text{ with } i \in [\![ 1, N]\!] \}$, where $(H, W)$ represents the input resolution, $C$ denotes the number of channels, $P \times P$ indicates the size of an image patch, and $N = \frac{H \times W}{P^2}$ denotes the number of patches. The ViT linearly maps each $\mathbf x_i$ into a patch embedding $\mathbf z^0_{i} \in \mathbb R^D$ using a learnable matrix $\mathbf E \in \mathbb R^{(C.P^2) \times D}$. Subsequently, a learnable embedding $\mathbf x_{cls} \in \mathbb R^D$ is prepended at the zero-index of embeddings $\mathbf z^0$ for the classification. Additionally, we use a learnable positional embedding $\mathbf E^{pos}$  to incorporate the position information of patches. The aforementioned process is described as follows,
\begin{equation}
    \mathbf z^0 = [\mathbf x_{cls}; \mathbf x_1\mathbf E; \mathbf x_2\mathbf E; \cdots; \mathbf x_N\mathbf E] + \mathbf E^{pos},
\label{equa:patch_embed}
\end{equation}
where $\mathbf E^{pos} \in \mathbb R^{(N+1) \times D}$ and $\mathbf z^0 \in \mathbb R^{(N+1) \times D}$. Afterward, $\mathbf z^0$ is fed into several transformer encoder blocks. Similar to ViT~\cite{ViT}, FakeFormer has $L$ blocks, each one containing a multi-head self-attention (MHSA), Layernorm (LN), and a multi-layer perceptron (MLP). The feature extraction process is described as follows,
% \begin{equation}
\begin{flalign}
    \mathbf z^l &= \text{MHSA}(\text{LN}(\mathbf z^{l - 1})) + \mathbf z^{l - 1}, \\\nonumber
    \mathbf z^{l'} &= \text{MLP}(\text{LN}(\mathbf z^l)) + \mathbf z^l, 
\end{flalign}
% \end{equation}
% with $l \in [\![ 1, L]\!]$ and $\mathbf z^l \in \mathbb R^{(N+1) \times D}$.
% \begin{equation}
%     \mathbf z^{l'} = \text{MLP}(\text{LN}(\mathbf z^l)) + \mathbf z^l,
% \end{equation}
with $l \in [\![ 1, L]\!]$ and $\mathbf z^l, \mathbf z^{l'} \in \mathbb R^{(N+1) \times D}$.
The extracted feature from the classification embedding $\mathbf z^{L'}_{0}$ is processed by a classification head composed of an MLP, resulting in the predicted category output $\hat{\mathbf y}$. In the task of deepfake detection, the categories consist of \textit{real} or \textit{fake}. Gray blocks in Figure~\ref{fig:fakeformer_ovv}a are parts of the used ViT.

\subsection{Attention to Vulnerable Patches}

To overcome the challenge discussed in Section~\ref{sec:investigation}, we propose integrating the Learning-based Local Attention Module (L2-Att) within ViT to guide the network to focus on the most artifact-prone patches, termed \textit{vulnerable patches}. In this section, we start by recalling the concept of blending-based data synthesis that is an essential component of L2-Att. Note that FakeFormer is trained using only real data and blending-based pseudo-fakes. Then, we define vulnerable patches, and detail their extraction process as well as the generation of the ground-truth data. Finally, we describe the architecture of the proposed L2-Att module.

\subsection{Background on Blending-based Data Synthesis}  This pseudo-fake generation technique has been widely used in deepfake detection~\cite{fxray,sbi}.
Given two real facial images $\mathbf I_B$ and $\mathbf I_F$, it consists in generating a manipulated image denoted as $\mathbf I_M$ as follows, 
\begin{equation}
    \mathbf I_M = \mathbf M \odot \mathbf I_F + (\mathbf 1 - \mathbf M) \odot \mathbf I_B \text{,}
    \label{equa:blending_formula}
\end{equation}
where $\odot$ denotes the element-wise multiplication, $\mathbf 1$ represents an all-one matrix, and $\mathbf M$ refers to the deformed convex hull.

\subsubsection{Vulnerable Patches}
\label{subsec:vulner_patch}

\begin{table*}
\centering
\scalebox{0.75}{
% \resizebox{\textwidth}{!}{
\begin{tabular}{c| c| c|c|c|c| H c}
\hline
$f_2$ & FF++ & CDF1 & CDF2 & DFD & DFW & DFDCP & DFDC \\
%  &  & AUC & AP & AR & mF1 & AUC & AP & AR & mF1 & AUC & AP & AR & mF1 & AUC & AP & AR & mF1 & AUC & AP & AR & mF1 & AUC & AP & AR & mF1 \\
 
\hline
\hline
mean & 97.45 & 96.29 & 94.16 & 95.03 & 81.02 & - & 78.28 \\
\cline{1-8}
max & \textbf{97.67(\textcolor{ForestGreen}{$\uparrow$0.22})} & \textbf{97.25(\textcolor{ForestGreen}{$\uparrow$0.96})} & \textbf{94.45(\textcolor{ForestGreen}{$\uparrow$0.29})} & \textbf{96.12(\textcolor{ForestGreen}{$\uparrow$1.09})} & \textbf{81.74(\textcolor{ForestGreen}{$\uparrow$0.72})} & \textbf{96.30} & \textbf{78.91(\textcolor{ForestGreen}{$\uparrow$0.63})} \\
\hline
\end{tabular}%
}
\vspace{-2mm}
\caption{\textbf{Selection of $f_2$}. AUC (\%) comparisons between different functions $f_2$ (Eq.~\eqref{equa:vulner_patch}). 
% The model is trained on FF++~\cite{ff++}.
% and evaluated on other datasets~\cite{celeb_df, dfd, wdf, dfdc}. 
}
\vspace{-4mm}
\label{tabl:vul_type_abl}
\end{table*}

Inspired by~\cite{laa_net}, we extend the definition of \textit{vulnerable points} to \textit{vulnerable patches}.
\begin{Definition}%[Fibration]
 - Vulnerable patches are the patches in an image that are the most likely to contain blending artifacts.
\end{Definition}
This definition allows tailoring the idea of local vulnerable regions to the self-attention mechanism in ViTs, which occurs at the patch level. As in~\cite{laa_net}, we propose focusing on blending artifacts that are generic across different deepfake generation methods, thereby helping to achieve good generalization capabilities to unseen deepfakes. Inspired by~\cite{laa_net}, as vulnerable patches represent the subregions that are the most impacted by the blending, we assume that they correspond to the blending regions with the highest mixture magnitude from the original images $\mathbf I_B $ and $\mathbf I_F$.
% LAA-Net~\cite{laa_net} leverages blending-based pseudo-fake synthesis algorithms~\cite{fxray, sbi} to identify vulnerable points where artifacts are likely to occur in the blending boundary. We adapt and extend this concept of vulnerable points into vulnerable patches, which are compatible with Transformer-based backbones. 
To generate vulnerable patches, we leverage blending-based synthesis techniques. We believe that data synthesis algorithms are convenient to ensure generalization while significantly reducing the need for large-scale training datasets, typically needed for transformers. Hence, similar to~\cite{fxray}, we calculate the blending boundary $\mathbf B=~({b}_{lm})_{l\in [\![ 1, H]\!], m  \in [\![ 1, W]\!]}$ resulting from the combination of $\mathbf I_B$ and $\mathbf I_F$ as follows,
\begin{equation}
    \mathbf B = 4 \cdot \mathbf M \odot (\mathbf 1 - \mathbf M) \text{,}
    \label{equa:blending_boundary}
\end{equation}
where $\odot$ denotes the element-wise multiplication and $\mathbf 1$ represents an all-one matrix. 
A higher value of $b_{rc}$ at a specific position $(r,c) \in [1, H] \times [1, W]$ indicates a more significant mixture between $\mathbf I_B$ and $\mathbf I_F$. We then apply to $\mathbf B$, a patching function  $f_1$  that extracts  $N$ non-overlapping patches denoted as $\Tilde{\mathbf B}=~(\Tilde{\mathbf B}_{ij})_{i, j \in [\![ 1, \sqrt{N}]\!]}$ of dimension $P \times P$  such that $\Tilde{\mathbf B}=f_1({\mathbf B})$. Finally, to quantify the vulnerability of each patch, a second function  $f_2$ operating at the patch level is defined. In summary, the vulnerable patches  $\mathcal P$ are obtained as follows,
\begin{equation}
    \mathcal P = \argmax_{(i,j)}(f_2(\Tilde{\mathbf B}_{ij})),
    \label{equa:vulner_patch}
\end{equation}
 For the sake of simplicity, we choose $f_2$ as the \textit{max} operation, which implies that $f=f_2 \circ f_1$ corresponds to maxpooling. We also explore the \textit{mean} operation in Section~\ref{subsec:ablation}. In future works, we intend to investigate additional patch-based vulnerability functions. Note that $\mathcal P$ can also include more than one patch, since $\Tilde{\mathbf B}$ can be maximal at several locations. This process is illustrated in Figure~\ref{fig:fakeformer_ovv}c.

\subsubsection{Ground Truth Generation for L2-Att} Finally, to obtain the ground truth to be compared to the output of L2-Att denoted as $\mathbf S$, we generate a weighted map $\mathbf S_{p}$ for each element $\mathbf p = (p_x, p_y) \in \mathcal P$. To take into account the neighborhood patches, we use an unnormalized Gaussian distribution to calculate $\mathbf S_{p}$ as follows,
\begin{equation}
    \mathbf S_{p}(x, y) = \exp\left(-\frac{(x - p_x)^2 + (y - p_y)^2}{2\sigma^2}\right),
    \label{equa:gauss_weighted_map}
\end{equation}
where $(x,y) \in [[1, \sqrt{N}]]$ represents the spatial position, and the standard deviation $\sigma$ is fixed to $1$ by default. We obtain $\mathbf S$ by overlaying $ \{\mathbf S_{p} \}_{p \in \mathcal P}$. The ground truth generation process is also illustrated in Figure~\ref{fig:fakeformer_ovv}c. It can be noted that, for real data, $\mathbf S$ is set to a zero matrix.

\subsubsection{Learning-based Local Attention Module (L2-Att) }
\label{subsec:lgam}
Our hypothesis is that since the patch size can be too large relative to the area of artifacts, the features encoded within a patch embedding may hold insufficient information about them. Consequently, the implicit self-attention mechanism might overlook or miss important patches, as those containing forgeries can appear too similar to those without. Therefore, we propose an explicit attention mechanism to ensure that the model pays more attention to these critical patches. To this end, by predicting the locations of vulnerable patches, L2-Att diffuses artifact-prone local information through the ViT. In particular, L2-Att first takes the patch embeddings $\mathbf z^{L'}_{\neg 0} \in \mathbb R^{N \times D}$ as input and processes them to produce spatial features as follows,
\begin{flalign}
    \mathbf F &= \text{Permute}(\mathbf z^{L'}_{\neg 0}), \hspace{3mm} \mathbf F \in \mathbb R^{D \times N} \text{,} \\\nonumber
    \mathbf F_{out} &= \text{Reshape}(\mathbf F), \hspace{1mm} \mathbf F_{out} \in \mathbb R^{D \times \sqrt{N} \times \sqrt{N}} \text{,}
    \label{equa:reshape_output}
\end{flalign}
% \begin{equation}
%     \mathbf F_{out} = \text{Reshape}(\mathbf F), \hspace{1mm} \mathbf F_{out} \in \mathbb R^{D \times \sqrt{N} \times \sqrt{N}} \text{,}
%     \label{equa:reshape_output}
% \end{equation}
% where the Permute operation changes the shape of $\mathbf z^{L'}_{\neg 0}$ from $N \times D$ to $D \times N$, while the Reshape further transforms it to $D \times \sqrt{N} \times \sqrt{N}$.

After that, $\mathbf F_{out}$ is fed into two convolution blocks (ConvBlock). The predicted weighted heatmap denoted as $\hat{\mathbf S}$ describing the presence probability of vulnerable patches is obtained as follows,
\begin{equation}
    \hat{\mathbf S} = \text{ConvBlock}_{1\times1}(\text{ConvBlock}_{3\times3}(\mathbf F_{out})),
    \label{equa:hm_output}
\end{equation}
where $\hat{\mathbf S} \in \mathbb R^{1 \times \sqrt{N} \times \sqrt{N}}$.
Each convolution block comprises a convolution layer, followed by a BatchNorm and a RELU layer. The notation $\text{ConvBlock}_{n\times n}$ indicates a convolution block with an ${n\times n}$ filter size. 
% Overall, L2-Att can provide robust spatial prior information about the location of vulnerable patches to the patch embeddings, complementing the self-attention mechanism of ViT. This prior information can also facilitate easier convergence, thereby reducing the amount of required training data for FakeFormer. 
A detailed illustration of the L2-Att  module can be seen in Figure~\ref{fig:fakeformer_ovv}b.

\begin{table}
% \vspace{-3mm}
\centering
% \scalebox{0.82}{
\resizebox{\linewidth}{!}{
\begin{tabular}{c| c| c| c| c| c| c }
\hline
\multirow{2}{*}{Method} & \multicolumn{2}{c|}{Training data} & \multirow{2}{*}{Params} & \multirow{2}{*}{FLOPs} & \multicolumn{2}{c}{Test set AUC(\%)} \\
 
\cline{2-3}\cline{6-7}
 & Real & Fake & & & FF++ & CDF2 \\
 
\hline
\hline
ViT-B~\cite{tall_swin} & $\checkmark$ & $\checkmark$ & 84M & 55.4G & - & 82.33 \\
% & 72.64 & - & - & - \\
ViT-B+TALL~\cite{tall_swin} & $\checkmark$ & $\checkmark$ & 84M & 55.4G & - & 86.58 \\
% & 74.10 & - & - & - \\
\hline
\hline

FakeFormer-S & $\checkmark$ & & 23M & 8.9G & 97.67 
% & 95.62 & 89.29 & 92.35 
& 94.45 \\
% & -- & - & - & - \\
FakeFormer-B & $\checkmark$ & & 91M & 35.8G & 97.76 
% & - & - & - 
& \underline{95.21} \\
% & \textbf{94.68} & - & - & - \\
\hline

\hline
Swin-B~\cite{tall_swin} & $\checkmark$ & $\checkmark$ & 86M & 47.5G & - & 83.13 \\
% & 73.01 & - & - & - \\
Swin-B+TALL~\cite{tall_swin} & $\checkmark$ & $\checkmark$ & 86M & 47.5G & 99.87 
% & - & - & - 
& 90.79 \\
% & 76.78 & - & - & - \\

\hline
\hline

% FakeSwin-T & 33M & 3.7G & 93.86 & - & - & - & - & - & - & - \\
FakeSwin-S & $\checkmark$ & & 55M & 6.5G & \underline{99.89} 
% & 99.76 & 96.42 & 98.07 
& 94.48 \\
% & - & - & - & - \\
FakeSwin-B & $\checkmark$ & & 91M & 11.5G & \textbf{99.94} 
% & - & - & - 
& \textbf{95.44} \\
% & \underline{93.55} & - & - & - \\
\hline
\end{tabular}%
}
\vspace{-2mm}
\caption{\textbf{Performance of Transformer-based approaches}. Comparisons in terms of AUC (\%) on CDF2~\cite{celeb_df}, and FF++~\cite{ff++}. Methods are trained on FF++~\cite{ff++} and tested on the other datasets. \textbf{Bold} and \underline{underlined} results highlight the best and the second-best performance, respectively.}
\vspace{-3mm}
\label{tabl:vit_auc}
\end{table}

\begin{table*}
\centering
\scalebox{0.77}{
% \resizebox{\textwidth}{!}{
\begin{tabular}{c| cc| ccHH|ccHH|ccHH|ccHH|ccHH|ccHH}
\hline
\multirow{3}{*}{Method} & \multicolumn{2}{c|}{Training set} & \multicolumn{24}{c}{Test set (\%)} \\
\cline{2-3}
\cline{4-27}

 & Real & Fake & \multicolumn{4}{c|}{CDF1} & \multicolumn{4}{c|}{CDF2} & \multicolumn{4}{c|}{DFW} & \multicolumn{4}{c|}{DFD} & \multicolumn{4}{c|}{DFDCP} & \multicolumn{4}{c}{DFDC} \\
 
\cline{4-27}
 & & & AUC & AP & AR & mF1 & AUC & AP & AR & mF1 & AUC & AP & AR & mF1 & AUC & AP & AR & mF1 & AUC & AP & AR & mF1 & AUC & AP & AR & mF1 \\
 
\hline
\hline
Xception~\cite{ff++} & $\checkmark$ & $\checkmark$ & 58.81 & 65.59 & 55.58 & 60.17 & 61.18 & 66.93 & 52.40 & 58.78 & 65.29 & 55.37 & 57.99 & 56.65 & 89.75 & 85.48 & 79.34 & 82.29 & 69.90 & 91.98 & 67.07 & 77.57 & 45.60 & - & - & - \\

FaceXRay~\cite{fxray} & $\checkmark$ & $\checkmark$ & 80.58 & 73.33 & - & - & - & - & - & - & - & - & - & - & 95.40 & 93.34 & - & - & 80.92 & 72.65 & - & - & - & - & - & - \\

% LRNet~\cite{lrnet} & $\checkmark$ & 52.84 & - & - & - & 53.20 & - & - & - & - & - & - & - & 52.29 & - & - & - & - & - & - & - & - & - & - & - \\

LocalRL~\cite{localRL} & $\checkmark$ & $\checkmark$ & - & - & - & - & 78.26 & - & - & - & - & - & - & - & 89.24 & - & - & - & 76.53 & - & - & - & - & - & - & - \\

% TI$^2$Net~\cite{ti2net} & $\checkmark$ & 66.65 & - & - & - & 68.22 & - & - & - & - & - & - & - & 72.03 & - & - & - & - & - & - & - & - & - & - & - \\

Multi-attentional~\cite{multi-attentional} & $\checkmark$ & $\checkmark$ & 69.14 & 74.03 & 52.70 & 61.57 & 68.26 & 75.25 & 52.40 & 61.78 & 73.56 & 73.79 & 63.38 & 68.19 & 92.95 & 96.51 & 60.76 & 74.57 & 83.81 & 96.52 & 77.68 & 86.08 & - & - & - & - \\

RECCE~\cite{ete_recons} & $\checkmark$ & $\checkmark$ & 49.96 & 63.04 & 50.87 & 56.31 
& 70.93 & 70.35 & 59.48 & 64.46 
& 68.16 & 54.41 & 56.59 & 55.48 
& \underline{98.26} & 79.42 & 69.57 & 74.17 
& 80.93 & 92.76 & 70.68 & 80.23 
& - & - & - & - \\

SFDG~\cite{sfdg} & $\checkmark$ & $\checkmark$ & - & - & - & - & 75.83 & - & - & - & 69.27 & - & - & - & 88.00 & - & - & - & 73.63 & - & - & - & - & - & - & - \\

EIC+IIE~\cite{eic_iie} & $\checkmark$ & $\checkmark$ & - & - & - & - & 83.80 & - & - & - & - & - & - & - & 93.92 & - & - & - & 81.23 & - & - & - & - & - & - & - \\

% AltFreezing~\cite{altfreezing} & $\checkmark$ & - & - & - & - & 89.50 & - & - & - & - & - & - & - & 98.50 & - & - & - & - & - & - & - & - & - & - & - \\

CADDM~\cite{caddm} & $\checkmark$ & $\checkmark$ & 89.36 & \underline{93.25} & \underline{81.41} & 86.93 & \underline{93.88} & \underline{91.12} & 77.00 & 83.46 & \underline{74.48} & \underline{75.23} & \underline{65.26} & \underline{69.89} & \textbf{99.03} & \textbf{99.59} & 82.17 & 90.04 & - & - & - & - & \underline{73.85} & - & - & - \\

UCF~\cite{ucf} & $\checkmark$ & $\checkmark$ & - & - & - & - & 82.4 & - & - & - & - & - & - & - & 94.5 & - & - & - & 80.5 & - & - & - & - & - & - & - \\

% Controllable GS~\cite{cgs} & $\checkmark$ & - & - & - & - & 84.97 & - & - & - & - & - & - & - & - & - & - & - & 81.65 & - & - & - & - & - & - & - \\

M2TR~\cite{m2tr} & $\checkmark$ & $\checkmark$ & - & - & - & - & 68.2 & - & - & - & - & - & - & - & - & - & - & - & - & - & - & - & - & - & - & - \\

DFDT~\cite{dfdt} & $\checkmark$ & $\checkmark$ & - & - & - & - & 88.3 & - & - & - & - & - & - & - & - & - & - & - & 76.1 & - & - & - & - & - & - & - \\

TALL-Swin~\cite{tall_swin} & $\checkmark$ & $\checkmark$ & - & - & - & - & 90.79 & - & - & - & - & - & - & - & - & - & - & - & 76.78 & - & - & - & - & - & - & - \\

LSDA~\cite{LSDA} & $\checkmark$ & $\checkmark$ & - & - & - & - & 91.10 & - & - & - & - & - & - & - & - & - & - & - & 77.00 & - & - & - & - & - & - & - \\

\hline
% FaceXRay+BI~\cite{fxray} &  & 74.76 & 68.99 & - & \multicolumn{1}{c|}{-} & - & - & - & \multicolumn{1}{c|}{-} & - & - & - & \multicolumn{1}{c|}{-} & 93.47 & 87.89 & - & - \\

% PCL+I2G~\cite{cstency_learning} &  & 98.30 & - & - & - & 90.03 & - & - & - & - & - & - & - & 99.07 & - & - & - & 74.27 & - & - & - & 67.52 & - & - & - \\

ICT~\cite{ict} & $\checkmark$ & & 81.43 & - & - & - & 85.71 & - & - & - & - & - & - & - & 84.13 & - & - & - & - & - & - & - & - & - & - & - \\

SBI~\cite{sbi} & $\checkmark$ & & \underline{92.53} & 79.91 & 79.16 & 79.53 
& 93.18 & 85.16 & \underline{82.68} & \underline{83.90} 
& 67.47 & 55.87 & 55.82 & 55.85 
& 97.56 & 92.79 & \underline{89.49} & 91.11 
& \underline{86.15} & \underline{93.24} & \underline{71.58} & \underline{80.99} 
& 72.42 & - & - & - \\

% AUNet~\cite{aunet} &  & - & - & - & - & 92.77 & - & - & - & - & - & - & - & \textbf{99.22} & - & - & - & \underline{86.16} & - & - & - & 73.82 & - & - & - \\

\hline
\hline
% Ours (w/ BI) &  & 91.67 & 94.79 & 50.0 & 65.47 & 86.28 & \underline{91.93} & 50.01 & \multicolumn{1}{c|}{64.78} & 57.13 & 56.89 & 50.12 & \multicolumn{1}{c|}{{53.29}} & \textbf{99.51} & \textbf{99.80} & \textbf{95.47} & \multicolumn{1}{c|}{\textbf{97.59}} & 69.69 & \underline{93.67} & 50.12 & 65.30 \\
% \hdashline

FakeFormer & $\checkmark$ & & \textbf{97.25} & \textbf{98.36} & 79.73 & 88.07 
& \textbf{94.45} & \textbf{97.15} & 81.29 & 88.51 
& \textbf{81.74} & \textbf{83.72} & 71.44 & 77.10 
& 96.12 & \underline{98.31} & 78.85 & 87.52 
& \textbf{96.30} & \textbf{99.50} & 78.01 & 87.45 
& \textbf{78.91} & \textbf{80.01} & 70.86 & 75.15 \\

% FakeFormer-B &  & \underline{97.56} & \underline{98.50} & 75.68 & 85.59
% & \underline{95.21} & \underline{97.57} & 77.92 & 86.64 
% &\underline{77.74} & \underline{80.95} & 67.57 & 73.66 
% & 96.61 & 98.65 & 77.45 & 86.77 
% & \textbf{94.97} & \textbf{99.30} & 75.91 & 86.04 
% & \underline{78.61} & \textbf{81.05} & 70.26 & 75.27 \\

% FakeSwin-S &  & 96.69 & 98.13 & 84.07 & 90.56 
% & 94.48 & 97.17 & 85.47 & 90.94 
% & 76.01 & 79.04 & 68.76 & 73.54 
% & 99.68 & 99.88 & 91.43 & 95.46 
% & 94.21 & 99.20 & 85.32 & 91.74 
% & 77.47 & 80.34 & 68.96 & 74.22 \\

% FakeSwin-B &  & \textbf{97.99} & \textbf{98.78} & 90.28 & 94.34 & \textbf{95.44} & \textbf{97.66} & 86.04 & 91.49 & \textbf{78.75} & \textbf{81.58} & 71.42 & 75.89 & \textbf{99.86} & \textbf{99.42} & 96.11 & 97.99 & \underline{93.55} & \underline{99.08} & 82.37 & 89.96 & \textbf{78.88} & \underline{80.97} & 69.70 & 75.07 \\
\hline
\end{tabular}%
}
\vspace{-2mm}
\caption{\textbf{Cross-dataset evaluation}. Comparisons in terms of AUC (\%) and AP (\%) on six benchmark datasets including CDF1~\cite{celeb_df}, CDF2~\cite{celeb_df}, DFW~\cite{wdf}, DFD~\cite{dfd}, DFDCP~\cite{dfdcp}, and DFDC~\cite{dfdc}. The results for comparisons are directly extracted from the original papers.}
\vspace{-3mm}
\label{tabl:cross_auc_full_metrics_suppl}
\end{table*}

\subsection{Training Objective}
\label{subsec:training_obj}

% \textbf{\textcolor{red}{Training objective:}}

To train FakeFormer, we optimize two losses, namely the Binary Cross Entropy (BCE) loss for classification denoted as $L_{cls}(\hat{\mathbf y}, \mathbf y)$, and the regression loss related to the prediction of vulnerable patches locations, denoted as $L_{att}(\hat{\mathbf S}, \mathbf S)$. Therefore, the total loss $L$ is defined as follows,
\begin{equation}
    L = L_{cls} + \lambda L_{att} \text{,}
    \label{equa:total_loss}
\end{equation}
where $\lambda$ is a balancing factor between the two losses and is empirically set to $10$. We employ the focal loss~\cite{focal_loss} to compute $L_{att}(\hat{\mathbf S}, \mathbf S)$. 

% \begin{equation}
%     \hat{\mathbf y} = \text{Sigmoid}(\text{FC}(\text{LN}(\mathbf z^{L'}_0)))
% \end{equation}
% where $(\hat{y}, y)$ correspond to the predictions and labels of 

% \subsection{Variants of Architecture}
% \label{subsec:variants}

% \textbf{\textcolor{red}{Architecture variants of FakeFormer:}}

% Besides the use of different synthesis algorithms~\cite{fxray, sbi}, we additionally provide two variants of FakeFormer. The FakeFormer is default referred to as the "Small" configuration, and the "Base" is mostly adopted from ViT-Base. These hyper-parameters of architecture variants are listed below:
% \begin{itemize}
%     \item FakeFormer(-S): $(H, W)=(112, 112)$, $(P,P)=(8,8)$, $L=12$, $D=384$, MLP size $=1536$, No. Heads $=6$, Params$=23$M, FLOPs$=8.9$G
%     \item FakeFormer(-B): $(H, W)=(224, 224)$, $(P,P)=(16,16)$, $L=12$, $D=768$, MLP size $=3072$, No. Heads $=12$, Params$=91$M, FLOPs$=35.8$G
% \end{itemize}

\section{Experiments}
\label{sec:experiment}

\subsection{Setups}
\label{subsec:setup}

% \textcolor{red}{This subsection covers Dataset + Data processing + Training + Evaluation metrics.}

\noindent\textbf{Datasets.} 
As in~\cite{fxray, sbi, lttd, ost, sladd, caddm, tall_swin}, we use the FaceForensics++ (FF++)~\cite{ff++} dataset for training.  Only real data and blended-based pseudo-fakes~\cite{fxray, sbi} are employed for training. We consider both the \textit{in-dataset} and the \textit{cross-dataset} protocols by testing on FF++ and other benchmarks including Celeb-DF (CDF1, CDF2)~\cite{celeb_df}, WildDeepfake (DFW)~\cite{wdf}, DeepFakeDetection (DFD)~\cite{dfd}, Deepfake Detection Challenge Preview (DFDCP)~\cite{dfdcp}, and Deepfake Detection Challenge (DFDC)~\cite{dfdc}, respectively.
% Additionally, besides the standard setup, we provide evaluation by combining all six datasets into a single dataset, referred to as the \textit{Combined} dataset. This combined dataset can better simulate real-world scenarios where various types of deepfakes are mixed. To the best of our knowledge, our work is the first to perform such an evaluation setup. 
More details about the datasets are provided in the supplementary materials.

\noindent\textbf{Data Processing.} Following the splitting convention~\cite{ff++}, we sample $128$, $32$, and $32$ frames from each video for training, validation, and testing respectively. This significantly reduces the amount of training data compared to previous ViT-based works~\cite{ict, dfdt, m2tr}. The facial regions are extracted using RetinaFace~\cite{retina_face}. To align two images for generating pseudo-fakes~\cite{fxray,sbi}, Dlib~\cite{dlib} is used as a landmark extractor. Further details are provided in the supplementary materials.

\noindent\textbf{Training.} We train FakeFormer for $200$ epochs using the AdamW~\cite{adamw} optimizer with a weight decay of $10^{-4}$ and a batch size of $32$. The weights are initialized using a ViT~\cite{dinov1, ViT} pretrained on ImageNet~\cite{imagenet}. The learning rate maintains at $5 \times 10^{-5}$ during the first quarter of iterations, then gradually decays to zero over the remaining epochs. We freeze the backbone (i.e., ViT without the head) for the first $6$ epochs, before training all layers. As a regularizer, we integrate LabelSmoothing~\cite{label_smoothing} with the loss function (Eq.~\ref{equa:total_loss}). 
For each video in the batch data, we dynamically select only $m$ frames, with $m=8$ or $m=16$ when using SBI~\cite{sbi} or BI~\cite{fxray}, respectively. 
Data augmentation techniques include color jittering, random cropping, scaling, horizontal flipping, Gaussian noise, blurring, and JPEG compression. All experiments are conducted on $4$ NVIDIA A100 GPUs.

\begin{table}
% \vspace{-3mm}
\centering
% \scalebox{0.77}{
\resizebox{\linewidth}{!}{
\begin{tabular}{c| c| c| c| c| c}
\hline
\multirow{2}{*}{Method} & \multirow{2}{*}{Params} & \multirow{2}{*}{FLOPs} & \multicolumn{3}{c}{Test set AUC(\%)} \\
\cline{4-6}
 & & 
 % & \multicolumn{1}{H}{FF++} 
 % & \multicolumn{4}{H}{CDF1} 
 & CDF2 
 % & \multicolumn{4}{H}{DFD} 
 & DFDCP & DFDC \\
 
% \cline{3-26}
%  &  & AUC & AP & AR & mF1 & AUC & AP & AR & mF1 & AUC & AP & AR & mF1 & AUC & AP & AR & mF1 & AUC & AP & AR & mF1 & AUC & AP & AR & mF1 \\
 
\hline
\hline
FaceXRay+BI~\cite{fxray} & 41.23M & -
% & \underline{98.52} 
% & 74.76 & - & - & - 
& 79.5 
% & \textbf{93.47} & - & - & - 
& 65.5
& - \\

% \hline
ICT~\cite{ict}+BI~\cite{fxray} & 21.45M & 8.4G
% & 90.22 
% & \underline{81.43} & - & - & - 
& 85.71 
% & 84.13 & - & - & - 
& - 
& - \\

LAA-Net~\cite{laa_net}+BI~\cite{fxray} & 27.13M & 11.6G
% & 99.95 
% & 92.46 & - & - & - 
& \underline{86.28} 
% & 99.51 & - & - & - 
& \underline{69.69}
& \underline{64.18} \\
% & 61.64 & 49.88 & 55.14 \\

\hline
\hline

FakeFormer+BI~\cite{fxray} & 22.77M & 8.9G
% & \textbf{99.13} 
% & \textbf{89.53} & - & - & - 
& \textbf{90.34}
% & \underline{93.42} & - & - & - 
& \textbf{78.71}
& \textbf{76.84} \\
\hline
\end{tabular}%
}
\vspace{-2mm}
\caption{\textbf{Performance using BI~\cite{fxray}}. Comparisons in terms of AUC (\%) on CDF2~\cite{celeb_df}, DFDCP~\cite{dfdcp}, and DFDC~\cite{dfdc}.}
\vspace{-2mm}
\label{tabl:bi_auc}
\end{table}

\noindent\textbf{Evaluation Metrics.} To compare FakeFormer with existing works, we employ two common evaluation metrics: the Area under the Curve (AUC) and the Average Precision (AP). For a fair comparison, similar to previous studies~\cite{caddm, cstency_learning, sbi, aunet, tall_swin, lipforensics}, we report the video-level performance. 
% Specifically, for the "Combined" dataset, we additionally provide two metrics: Average Recall (AR) and mean F1-score (mF1), to offer a more comprehensive evaluation.

\begin{table}
% \vspace{-3mm}
\centering
% \scalebox{0.8}{
\resizebox{\linewidth}{!}{
\begin{tabular}{c| c| c| c| c| c}
\hline
\multirow{2}{*}{Method} & \multirow{2}{*}{Params} & \multirow{2}{*}{FLOPs} & \multicolumn{3}{c}{Test set AUC(\%)} \\
\cline{4-6}
 & & & CDF2 & DFDC & DFDCP \\
 % & \multicolumn{4}{H}{DFDC} \\

\hline
\hline
SBI~\cite{sbi} & 19.34M & 8.3G & 93.18 & 72.42 & 86.15 \\
% & - & - & - & - \\

LAA-Net~\cite{laa_net}+SBI~\cite{sbi} & 27.13M & 11.6G & \textbf{95.40} & \underline{72.43} & \underline{86.94} \\
% & - & - & - & - \\

% \hline
% CADDM+SBI~\cite{caddm} & $\checkmark$ & $\checkmark$ & 289.74M & 9.6G & 94.15 & \textbf{79.57} & - \\
% % & - & - & - & - \\

\hline
\hline
FakeFormer+SBI~\cite{sbi} & 22.77M & 8.9G & \underline{94.45} & \textbf{78.91} & \textbf{96.30}\\
% & \textbf{76.84} & - & - & - \\
\hline
\end{tabular}%
}
\vspace{-2mm}
\caption{\textbf{Performance using SBI~\cite{sbi}}. Comparisons in terms of AUC (\%) on CDF2~\cite{celeb_df}, DFDCP~\cite{dfdcp}, and DFDC~\cite{dfdc}. 
% Models are trained on FF++~\cite{ff++} and tested on unseen datasets. 
}
\vspace{-3mm}
\label{tabl:sbi_auc}
\end{table}

\noindent\textbf{Architectures.}
We utilize two variants of FakeFormer that we call FakeFormer-S and FakeFormer-B. By default, we use the lightweight FakeFormer-S, where $H=W=112$ and $P=8$. FakeFormer is based on the vanilla vision transformer, i.e., ViT~\cite{ViT}. However, although FakeFormer is based on ViT, we also assess its applicability using another transformer architecture, namely \textit{FakeSwin}, which is based on Swin~\cite{swin}. Similarly to FakeFormer, we consider two variants: FakeSwin-S and FakeSwin-B. Further details are provided in the supplementary materials.

\begin{table}
\centering
%\setlength\tabcolsep{2pt}
% \scalebox{0.72}{
\resizebox{\columnwidth}{!}{%
\begin{tabular}{c|cc|cccccH|c}
\hline
{Method} & Real & {Fake} & Saturation & Contrast & Block & Noise & Blur & Pixel & Avg \\ 
\hline
\hline
{FaceXray~\cite{fxray}} & {$\checkmark$} & {$\checkmark$} & 97.6 & 88.5 & \underline{99.1} & 49.8 & 63.8 & 88.6 & 79.8 \\

{LipForensics~\cite{lipforensics}} & {$\checkmark$} & {$\checkmark$} & \textbf{99.9} & 99.6 & 87.4 & 73.8 & \underline{96.1} & 95.6 & \underline{91.4} \\

{LAA-Net~\cite{laa_net}} & {$\checkmark$} & & \textbf{99.9} & \textbf{99.9} & \textbf{99.9} & 53.9 & \textbf{98.2} & 99.80 & 90.4 \\
\hline
\hline

FakeFormer & {$\checkmark$} & & 98.1 & 96.0 & 97.0 & \underline{75.3} & 87.4 & 91.2 & 90.8 \\
FakeSwin & {$\checkmark$} & & \underline{99.8} & \underline{99.8} & \textbf{99.9} & \textbf{81.6} & 89.8 & 93.7 & \textbf{94.2} \\
\hline
\end{tabular}%
}
\vspace{-2mm}
\caption{\textbf{Robustness to unseen perturbations}. Focusing on vulnerable patches, FakeFormer shows improved robustness to common perturbations, especially to noise.}
\vspace{-3mm}
\label{tabl:ff_noise_auc}
\end{table}

% \subsection{Generalization of LGAM and Vulnerable Patches to other Transformers}
% \label{subsec:fakeswin}

% % \textcolor{red}{This subsection introduce a combination of Swin + LGAM.}

% % \begin{itemize}
% %     \item \textcolor{red}{The purpose is to give more evidence to the generic of LGAM + vulnerable patch. }
% %     \item \textcolor{red}{At the same time, discussing that more powerful backbone to capture local artifacts => better overall performance in DF detection.}
% %     \item \textbf{\textcolor{red}{I am not really sure whether is it suitable to put the section here or other place.}}
% % \end{itemize}

% We further provide an additional integration of LGAM into a state-of-the-art Transformer backbone, Swin~\cite{swin}. we name the architecture, ``FakeSwin'', for later reference. The purpose of the architecture is to provide a more comprehensive analysis of the compatibility of the proposed vulnerable patches and LGAM to Transformer-based backbones. On the other side, we would provide more evidence of the relationship between the overall performance and the local feature extraction capacity of integrated backbones. Indeed, Swin~\cite{swin} is designed for diverse vision tasks, such as classification, dense predictions, etc. Based on Swin, we go with two variants of FakeSwin, the "Small" and the "Base" configuration. 

\subsection{Comparison with State-of-the-art}

% \textcolor{red}{This section for comparisons:}

% \begin{itemize}
%     \item \textcolor{red}{Compare between Transformer-based approaches}
%     \item \textcolor{red}{Compare between synthesis-based approaches}
%     \item \textcolor{red}{Compare in real-life environment}
% \end{itemize}

In this section, we compare the performance of FakeFormer to SOTA methods. In all tables, \textbf{bold} and \underline{underlined} results highlight the best and the second-best performance, respectively.
% We compare it to transformer-based methods in Section~\ref{subsec:in_dataset_eval} and approaches that utilize blending-based pseudo-fake generation in Section~\ref{subsec:cross_dataset_eval}. Furthermore, in Sections~\ref{subsec:cross_dataset_eval}, we compare it to the SoA methods in the cross-dataset setting.
% Furthermore, in Sections~\ref{subsec:cross_dataset_eval} and \ref{subsec:real_life_eval}, we compare it to the SoA in the cross-dataset setting and on the Combined dataset, respectively.

\begin{table*}[]
\centering
\scalebox{0.7}{
% \resizebox{\textwidth}{!}{
\begin{tabular}{c| c| Hc| Hc|Hc|Hc|Hc|c}
\hline
\multirow{1}{*}{Model} & \multirow{1}{*}{L2-Att} & \multicolumn{2}{c|}{FF++} & \multicolumn{2}{c|}{CDF1} & \multicolumn{2}{c|}{CDF2} & \multicolumn{2}{c|}{DFD} & \multicolumn{2}{c|}{DFDC} & Avg. \\
% \cline{3-12}
 % & & BI & SBI & BI & SBI & BI & SBI & BI & SBI & BI & SBI \\

% \cline{3-26}
%  &  & AUC & AP & AR & mF1 & AUC & AP & AR & mF1 & AUC & AP & AR & mF1 & AUC & AP & AR & mF1 & AUC & AP & AR & mF1 & AUC & AP & AR & mF1 \\
 
\hline
\hline
ViT~\cite{ViT}  & & - & 97.48 
& 85.18 & 94.42 
& 83.96 & 92.62 
& 92.71 & 95.72 
& 77.65 & 77.35 
& 91.52 \\
FakeFormer & $\checkmark$ & - & \textbf{97.67(\textcolor{ForestGreen}{$\uparrow$0.19})} 
 & \textbf{89.53(\textcolor{ForestGreen}{$\uparrow$4.35})} & \textbf{97.25(\textcolor{ForestGreen}{$\uparrow$2.83})} 
 & \textbf{90.34(\textcolor{ForestGreen}{$\uparrow$6.38})} & \textbf{94.45(\textcolor{ForestGreen}{$\uparrow$1.83})} 
 & \textbf{93.42(\textcolor{ForestGreen}{$\uparrow$0.71})} & \textbf{96.12(\textcolor{ForestGreen}{$\uparrow$0.40})} 
 & \textbf{78.71(\textcolor{ForestGreen}{$\uparrow$1.06})} & \textbf{78.91(\textcolor{ForestGreen}{$\uparrow$1.56})} 
 & \textbf{92.88(\textcolor{ForestGreen}{$\uparrow$1.36})} \\
\hline
\hline

% 98.94 is FakeFormer SBI results

Swin~\cite{swin} & & - & 99.75 
& 71.23 & 94.46 
& 82.50 & 90.89 
& 99.89 & 99.59 
& \textbf{69.94} & 74.20 
& 91.78 \\
% & 77.23 \\
FakeSwin & $\checkmark$ & - & \textbf{99.89(\textcolor{ForestGreen}{$\uparrow$0.14})} 
 & \textbf{75.85(\textcolor{ForestGreen}{$\uparrow$4.62})} & \textbf{96.69(\textcolor{ForestGreen}{$\uparrow$2.23})} 
 & \textbf{83.29(\textcolor{ForestGreen}{$\uparrow$0.79})} & \textbf{94.48(\textcolor{ForestGreen}{$\uparrow$3.59})} 
 & \textbf{99.94(\textcolor{ForestGreen}{$\uparrow$)0.05}} & \textbf{99.68(\textcolor{ForestGreen}{$\uparrow$0.09})} 
 & 68.48(\textcolor{ForestGreen}{$\downarrow$1.46}) & \textbf{77.47(\textcolor{ForestGreen}{$\uparrow$3.27})} 
 & \textbf{93.64(\textcolor{ForestGreen}{$\uparrow$1.86})} \\
\hline
\end{tabular}%
}
\vspace{-2mm}
\caption{\textbf{L2-Att ablation.} Performance of FakeFormer without and with L2-Att using the cross-dataset setting.}
\vspace{-3mm}
\label{tabl:lga_abl}
\end{table*}

% \subsubsection{Comparisons with Transformer-based Approaches}
% \label{subsec:in_dataset_eval}
\noindent\textbf{Comparison with Transformer-based Approaches.} Table~\ref{tabl:vit_auc} compares FakeFormer to Transformer-based methods in terms of model size (Params), computational cost (FLOPs), and AUC on FF++ and CDF2. Even with the smaller variant (i.e., -S), our method outperforms vanilla ViT-B and Swin-B. This highlights the effectiveness of the proposed L2-Att module, resulting in a lightweight model. Moreover, regardless of the transformer backbone, FakeFormer and FakeSwin generalize better than TALL~\cite{tall_swin} despite their simplicity. It should be noted that the ViT-B and the Swin-B results are directly recovered from \cite{tall_swin}, which means that the backbone configurations might be different.

\begin{table*}[]
\centering
\scalebox{0.75}{
% \resizebox{\textwidth}{!}{
\begin{tabular}{c| c| c| c| Hcccc|c}
\hline
\multirow{2}{*}{Model} & \multirow{2}{*}{Target} & \multirow{2}{*}{Params} & \multirow{2}{*}{FLOPs} & \multicolumn{6}{c}{Test set AUC(\%)}\\
\cline{5-10}
 & & & & FF++ & CDF1 & CDF2 & DFD & DFDC & Avg. \\

% \cline{3-26}
%  &  & AUC & AP & AR & mF1 & AUC & AP & AR & mF1 & AUC & AP & AR & mF1 & AUC & AP & AR & mF1 & AUC & AP & AR & mF1 & AUC & AP & AR & mF1 \\
 
\hline
\hline
\multirow{2}{*}{FakeFormer} & VPoint & 23.61M & 9.4G & - & 97.22 & 93.39 & 93.70 & 77.71 & 90.51 \\
 & VPatch & 22.77M & 8.9G & - & \textbf{97.25} & \textbf{94.45} & \textbf{96.12} & \textbf{78.91} & \textbf{91.68(\textcolor{ForestGreen}{$\uparrow$1.17})} \\
\hline
\hline

% 98.94 is FakeFormer SBI results

\multirow{2}{*}{FakeSwin} & VPoint & 57.25M & 8.1G & - & 96.16 & 94.25 & 99.30 & 74.99 & 91.18 \\
% & 77.23 \\
 & VPatch & 54.89M & 6.5G & - & \textbf{96.69} & \textbf{94.48} & \textbf{99.68} & \textbf{77.47} & \textbf{92.08(\textcolor{ForestGreen}{$\uparrow$0.9})} \\
\hline
\end{tabular}%
}
\vspace{-2mm}
\caption{\textbf{VPoint versus VPatch.} AUC (\%) when using Vulnerable Points~\cite{laa_net} (VPoint) versus Vulnerable Patches (VPatch).}
\vspace{-4mm}
\label{tabl:vpart_vpoint_abl}
\end{table*}

% \subsubsection{Comparisons Based on the Blending-based Synthesis Approaches}
% \label{subsec:cross_dataset_eval}

\noindent\textbf{Comparison with Blending-based Synthesis Approaches.} Table~\ref{tabl:bi_auc} and Table~\ref{tabl:sbi_auc} compare our method to SOTA when using Blending-Images (BI)~\cite{fxray} and Self-Blended-Images (SBI)~\cite{sbi}, respectively. As observed, we generally achieve better performances than SOTA methods using the same blending algorithm on CDF2, DFDC, and DFDCP datasets, despite training on a drastically lower number of data as compared to other ViT-based approaches such as ICT~\cite{ict}. Moreover, LAA-Net~\cite{laa_net}, as a CNN-based method, requires a larger model size as compared to our lightweight FakeFormer.

% \subsubsection{Overall Cross-dataset Evaluation}
% \label{subsec:cross_dataset_eval}

\noindent\textbf{Overall Cross-dataset Evaluation.} The overall comparison with SOTA methods under the cross-dataset setup can be seen in Table~\ref{tabl:cross_auc_full_metrics_suppl}. As observed, FakeFormer generalizes better than SOTA methods, including those trained with fake data, those trained with pseudo-fakes (without fake data), and transformer-based models such as ICT~\cite{ict}, M2TR~\cite{m2tr}, and DFDT~\cite{dfdt}, although our model is trained with a significantly lower amount of data.

\begin{table}
% \vspace{-3mm}
\centering
\scalebox{0.77}{
% \resizebox{\textwidth}{!}{
\begin{tabular}{c| H c| c| c| c| c}
\hline
$\lambda$ & FF++ & CDF1 & CDF2 & DFD & DFDC & Avg. \\
 
% \cline{3-26}
%  &  & AUC & AP & AR & mF1 & AUC & AP & AR & mF1 & AUC & AP & AR & mF1 & AUC & AP & AR & mF1 & AUC & AP & AR & mF1 & AUC & AP & AR & mF1 \\
 
\hline
\hline
1 & - & 94.68 & 93.67 & 94.62 & 77.91 & 90.22 \\
\cline{1-7}
10 & - & 97.25 & 94.45 & \textbf{96.12} & \textbf{78.91} & \textbf{91.68} \\
\cline{1-7}
100 & - & \textbf{97.86} & \textbf{94.96} & 94.88 & 78.58 & 91.57 \\
\hline
\end{tabular}%
}
\vspace{-2mm}
\caption{\textbf{Variation of the parameter $\lambda$}. AUC (\%) comparison using different $\lambda$ values (Eq.~\eqref{equa:total_loss}).}
\vspace{-4mm}
\label{tabl:lga_impact_abl}
\end{table}

\noindent\textbf{Robustness to Unseen Perturbations.} Since deepfakes can be altered by common perturbations on social media. Following the evaluation setting of ~\cite{DFo}, experiments evaluating the robustness of FakeFormer against five types of perturbation on FF++~\cite{ff++} are reported in Table~\ref{tabl:ff_noise_auc}. We can observe the superiority of FakeFormer and FakeSwin as compared to existing methods. 
\subsection{Ablation Study}
\label{subsec:ablation}

% \textcolor{red}{This section for ablation study:}

% \begin{itemize}
%     \item \textcolor{red}{Ablation study of the algorithm to construct vulnerable patches.~\ref{tabl:vul_type_abl}.}
%     \item \textcolor{red}{Ablation study on LGAM.}
%     \item \textcolor{red}{Ablation study on between using Vpoints and Vpatches.}
%     \item \textcolor{red}{Ablation study on the balancing factor $\lambda$ in Eq.~\ref{equa:total_loss}.}
% \end{itemize}

\noindent\textbf{Impact of L2-Att.} To validate the L2-Att module, 
we compare ViT~\cite{ViT}/Swin~\cite{swin} and FakeFormer (ViT+L2-Att)/FakeSwin (Swin+L2-Att) (with SBI~\cite{sbi}). The results on several datasets~\cite{ff++, celeb_df, dfd, dfdc} are presented in Table~\ref{tabl:lga_abl}. In general, L2-Att consistently contributes to the enhancement of both ViT and Swin, confirming the relevance of the proposed explicit attention mechanism. 
% However, interestingly, a slight drop in performance is observed when using L2-Att on DFDC for the case of Swin with BI~\cite{fxray}. Nevertheless, L2-Att performs well for all datasets when using a more subtle data synthesis, namely, SBI~\cite{sbi}.

\noindent\textbf{Vulnerable Points versus Vulnerable Patches.} 
LAA-Net~\cite{laa_net} proposes an attention mechanism based on vulnerable points (VPoint). To demonstrate the compatibility of vulnerable patches (VPatch) with transformers, we report in Table~\ref{tabl:vpart_vpoint_abl} the obtained results when replacing VPatch with VPoint within FakeFormer and FakeSwin. We note that the use of VPatch not only results in better performance but also guarantees a lower computational cost as compared to VPoint. We note that the higher number of parameters and FLOPs associated with using VPoint is caused by the decoder designed to locate VPoint. Meanwhile, VPatch does not require any decoder, making it more computationally efficient.
% To compare the suitability between vulnerable points (i.e., heatmap ground truth proposed in ~\cite{laa_net}) (VPoint) and vulnerable patches (VPatch), we follow the setup~\cite{laa_net} to train FakeFormer and FakeSwin by using the VPoint as the ground-truths. The experiment results are reported in Table~\ref{tabl:vpart_vpoint_abl}. Our experiment shows that the use of VPatch not only gains better results but also helps reduce computational costs as compared to VPoint. We note that the higher number of parameters and FLOPs of using VPoint is caused by the decoder designed to locate VPoint. Meanwhile, our VPatch does not require any decoder for the task; thus, more computationally efficient. 

\noindent\textbf{Selection of $f_2$.} Table~\ref{tabl:vul_type_abl} compares two aggregation functions $f_2$ defined in Eq.~\eqref{equa:vulner_patch} coupled with FakeFormer: \textit{mean} and \textit{max} operations. In both cases, the stability of the results can be seen.  By default, we select the \textit{max} operation as it gives slightly better results.

\noindent\textbf{Analysis of Parameter $\lambda$.} The weight $\lambda$ defined in  Eq.~\eqref{equa:total_loss} is set empirically to $10$ as it yields the best performance on average. We report the results using different values of $\lambda$ within FakeFormer in Table~\ref{tabl:lga_impact_abl}. It can be observed the generalization across four testing benchmarks remains robust regardless of the value of $\lambda$.

\section{Limitations}
\label{sec:limitation}

% FakeFormer use vulnerable-patch based attention based on blending-based synthesis algorithm that need 2 images to generate the pseudo-fake. Therefore, testing on fully synthesis data such as stylegan we got only 50.84\% AUC and 59.17\% AUC with FakeFormer and FakeSwin, respectively. Note that we only do not use StyleGAN2 to train our model. Additionally, we similar to other image-based methods, our method cannot leverage temporal info to deepfake videos.

As FakeFormer is tailored to blending-based manipulations such as face-swaps and face reenactments, it shows poor performance when tested on fully synthetic data. Indeed, we obtain $50.84$\% and $59.17$\% AUC when testing FakeFormer and FakeSwin on $1000$ randomly generated images using stylegan2~\cite{stylegan2}. 

% GAN images: FakeFormer: 50.84 AUC, FakeSwin: 59.17 AUC.

%FF++ c23, FakeFormer: 94.47 AUC, FakeSwin: 96.01 AUC.

\section{Conclusion}
\label{sec:conclu}

In this paper, we identify the shortcomings of ViTs in capturing localized features, explaining its restricted adoption in the field of deepfake detection. To address this, we introduce FakeFormer, a lightweight ViT-based framework for generalizable deepfake detection. FakeFormer incorporates L2-Att, an attention strategy directing the focus toward local inconsistency-prone regions, thus enhancing the detection capabilities of transformers. Leveraging the synthetic data generation and the proposed attention mechanism mitigates the need for extensive training data and large models. Detailed experiments demonstrate the superiority of FakeFormer over state-of-the-art methods. In future work, we will investigate strategies to extend FakeFormer for capturing temporal information and imposing more robustness to fully synthetic deepfakes.

% \clearpage
\newpage
% \centering
% \Large
% ---Supplementary Material---

\section{Appendix}

\subsection{Code Release}
The code is provided as a part of the supplementary materials. The code is under license registration and will be publicly released soon with pretrained weights.

\begin{table*}
\centering
\scalebox{0.9}{
% \resizebox{\textwidth}{!}{
\begin{tabular}{c| c|c|c|c|c|c}
\hline
\multirow{2}{*}{Method} & \multicolumn{6}{c}{Mask SSIM Ranges} \\
\cline{2-7}
 & $0.6$-$0.8$ & $0.8$-$0.9$ & $0.9$-$0.925$ & $0.925$-$0.95$ & $0.95$-$0.97$ & $0.97$-$1.0$ $\uparrow$ \\
 
\hline
\hline
ViT+SBI~\cite{sbi} & 90.5 & 94.0 & 89.2 & 84.7 & 82.1 & 78.0 \\
\cline{1-7}
EFNB4+SBI~\cite{sbi} & 89.3 & 84.1 & 80.3 & 83.6 & 82.8 & 78.6 \\
\cline{1-7}
Xception+SBI~\cite{sbi} & 82.9 & 80.6 & 80.5 & 85.0 & 82.3 & 79.4 \\
\cline{1-7}
CADDM~\cite{caddm} & 84.1 & 85.9 & 82.8 & 85.4 & 84.0 & 81.9 \\
\cline{1-7}
Swin+SBI~\cite{sbi} & 90.0 & 91.0 & 89.5 & 85.6 & 85.8 & 85.7 \\
\cline{1-7}
LAA-Net~\cite{laa_net} & 96.1 & 94.3 & 92.9 & 86.0 & 93.4 & 91.1 \\
\hline
\end{tabular}%
}
% \vspace{-2mm}
\caption{\textbf{Performance analyses across various levels of deepfake quality}. 
AUC (\%) comparisons between CNN-based methods (EFNB4~\cite{efn_net}+SBI~\cite{sbi}, Xception~\cite{xception}+SBI~\cite{sbi}, CADDM~\cite{caddm}, LAA-Net+SBI~\cite{laa_net}) and ViT-based methods (ViT~\cite{ViT}+SBI~\cite{sbi}, Swin~\cite{swin}+SBI~\cite{sbi}) across different deepfake quality levels assessed by the Mask SSIM~\cite{mssim_pose} metric.
The model is trained on FF++~\cite{ff++} and evaluated on CDF2~\cite{celeb_df}. 
}
% \vspace{-4mm}
\label{tabl:cnns_vit_ssim_abl}
\end{table*}

\begin{table*}
\centering
\scalebox{0.9}{
% \resizebox{\textwidth}{!}{
\begin{tabular}{c| c| c|c|c|c| c| c}
\hline
Input\&Patch Size & FF++ & CDF1 & CDF2 & DFD & DFW & DFDCP & DFDC \\
%  &  & AUC & AP & AR & mF1 & AUC & AP & AR & mF1 & AUC & AP & AR & mF1 & AUC & AP & AR & mF1 & AUC & AP & AR & mF1 & AUC & AP & AR & mF1 \\
 
\hline
\hline
$112$P$16$ & 81.83 & 87.93 & 85.26 & 73.56 & 76.87 & 93.28 & 72.53 \\
\cline{1-8}
$112$P$8$ & 97.67 & 97.25 & 94.45 & 96.12 & 81.74 & 96.30 & 78.91 \\
\cline{1-8}
$224$P$8$ & \textbf{99.93} & \textbf{99.00} & \textbf{96.84} & \textbf{99.54} & \textbf{82.11} & \textbf{96.99} & \textbf{79.01} \\
\hline
\end{tabular}%
}
% \vspace{-2mm}
\caption{\textbf{Effect of patch size}. AUC (\%) comparisons are presented for different setups of input resolution and patch size. The model is trained on FF++~\cite{ff++} and evaluated on other datasets~\cite{celeb_df, dfd, wdf, dfdc, dfdcp}. 
}
% \vspace{-4mm}
\label{tabl:patch_size_abl}
\end{table*}

\subsection{More Details regarding the Datasets}
\label{subsec:dataset_supp}

\noindent\textbf{Datasets.} FF++ contains $1000$ original videos and $4000$ deepfake videos. The types of deepfakes include Deepfakes (DF)~\cite{deepfake1}, Face2Face (F2F)~\cite{face2face}, FaceSwap (FS)~\cite{faceswap}, and NeuralTextures (NT)~\cite{neutex}. Moreover, while CDF and DFD represent datasets including high-quality deepfakes~\cite{celeb_df, laa_net}, DFW stands out as one of the most challenging benchmarks. DFW is collected entirely from the internet without any prior knowledge regarding the used manipulation techniques. Additionally, it includes images with occlusions, illumination changes, different head poses, and diverse backgrounds. Meanwhile, DFDC, including its preview version called DFDCP, comprises more than $128,000$ videos and is therefore one of the largest datasets in the field of deepfake detection.

\noindent\textbf{Data Processing.} 
The RetinaFace~\cite{retina_face} algorithm is employed to extract facial bounding boxes. These boxes are enlarged by a factor of $1.25$ and then resized to a specific resolution $(H, W)$ to match the different backbone configurations. To generate pseudo-fakes, the two images are first aligned. For this purpose, the Dlib~\cite{dlib} library is utilized to extract $68$ and $81$ facial landmarks for the two considered blending synthesis algorithms, BI~\cite{fxray} and SBI~\cite{sbi}, respectively. Finally, the processed pseudo-fakes are generated for each training iteration.
% The RetinaFace~\cite{retina_face} is adopted to extract bounding boxes of facial regions. These boxes are conservatively enlarged by a factor of $1.25$, then resized to a specific resolution $(H, W)$ corresponding to different training configs. We align images before applying algorithms to generate pseudo-fakes. For this purpose, the Dlib~\cite{dlib} is utilized to extract facial landmarks with $68$ and $81$ keypoints respective to two considered BI~\cite{fxray}, and SBI~\cite{sbi} blending synthesis algorithms. Finally, the processed images and the meta-data are stored for online pseudo-fake generation in the training process.

\subsection{Architecture Details}
We describe in detail the hyperparameters of the two considered FakeFormer variants as follows:
\begin{itemize}
    \item FakeFormer-S: $H=W=112$, $P=8$, $L=12$, $D=384$, MLP size $=1536$, No. Heads $=6$, Params$=23$M, FLOPs$=8.9$G.
    \item FakeFormer-B: $H= W=224$, $P=16$, $L=12$, $D=768$, MLP size $=3072$, No. Heads $=12$, Params$=91$M, FLOPs$=35.8$G.
\end{itemize}
where the \textit{MLP size} represents the dimension of hidden layers in MLP, the \textit{No. Heads} denotes the number of heads in MHSA, \textit{Params} is the number of parameters, and \textit{FLOPs} represents the computational cost in terms of floating point operations per second. For FakeSwin architecture, we adopt these two backbone variants from Swin~\cite{swin}, namely:
\begin{itemize}
    \item FakeSwin-S: $H=W=224$, $P=4$, $M=7$, $d=32$, $\alpha=4$, $C_h=96$, Layer Numbers = \{2, 2, 18, 2\}, No. Heads = \{3, 6, 12, 24\}, Params$=55$M, FLOPs$=6.5$G.
    \item FakeSwin-B: $H=W=224$, $P=4$, $M=7$, $d=32$, $\alpha=4$, $C_h=128$, Layer Numbers = \{2, 2, 18, 2\}, No. Heads = \{4, 8, 16, 32\}, Params$=91$M, FLOPs$=11.5$G.
\end{itemize}
where $M$ is the window size, $d$ is the query dimension of each head, the expansion layer of each MLP is $\alpha$, and $C_h$ denotes the channel number in the hidden layers during the first stage.
% As for the other configurations of FakeSwin, we follow~\cite{swin}.

\subsection{More Details regarding the training Setup in Section~\textcolor{red}{3}}

In this section, we present more details related to the experimental settings in Section~\textcolor{red}{3} of the main paper. 

Specifically, in Figure~\textcolor{red}{3}-a, we train vanilla ViT~\cite{ViT} and Swin~\cite{swin} with SBI data synthesis~\cite{sbi} using the same training settings described in~\cite{laa_net, sbi}. In addition, the detection performance of three CNN methods in different ranges of Mask-SSIM~\cite{mssim_pose} on CDF2~\cite{celeb_df}, i.e., LAA-Net~\cite{laa_net}, CADDM~\cite{caddm}, and SBI~\cite{sbi} for the comparison is directly extracted from~\cite{laa_net}. Note that, in~\cite{sbi}, the authors focus solely on data synthesis. Hence, the used architecture is a standard CNN (EfficientNet-B4 (EFNB4)~\cite{efn_net}). 
% Therefore, the comparison between the ViT and Swin experiments (also trained with SBI data synthesis) to it is fair.

Secondly, in Figure.~\textcolor{red}{3}-b, all models including CNNs and variants of ViTs are trained on FF++~\cite{ff++} with both real and fake data for $50$ epochs. Following conventional spitting~\cite{ff++}, we uniformly extract $128$ and $32$ frames of each video for training and validation, respectively. Hence, there are in total of $460800$ and $22400$ frames for the corresponding task. The weights of models are initialized by pretrained on ImageNet~\cite{imagenet}. We employ different optimizers as Adam is often used with CNNs~\cite{sladd, ost, caddm, fxray, multi-attentional, sfdg} and AdamW with ViTs~\cite{swin, twin, deit, PVT}. The learning rate is initially set to $10^{-4}$ and linearly decays to $0$ at the end of the training period. All experiments are carried out using a NVIDIA A100 GPU.

\subsection{Additional Results and Analysis related to Section~\textcolor{red}{3}}

In the main paper, Figure.~\textcolor{red}{3}-a shows the performance comparison of transformer-based approaches (ViT~\cite{ViT}+SBI~\cite{sbi} and Swin~\cite{swin}+SBI~\cite{sbi}) and CNN-based methods (EFNB4+SBI~\cite{sbi}, CADDM~\cite{caddm}, LAA-Net~\cite{laa_net}+SBI~\cite{sbi}) across different ranges of Mask-SSIM~\cite{mssim_pose}. Even though the three considered CNN methods are based on EFNB4~\cite{efn_net}, most of these methods have been specifically tailored for deepfake detection. Therefore, to support our claims. in addition to EFNB4~\cite{efn_net}, we further examine the performance of XceptionNet~\cite{xception}+SBI~\cite{sbi} on those Mask SSIM ranges under the same comparison setting. Besides EFNB4, Xception is one of the most commonly used backbones in deepfake detection~\cite{ff++, sbi, sladd, aunet, ucf}. The overall comparison including the results reported in Figure.~\textcolor{red}{3}-b is given in Table.~\ref{tabl:cnns_vit_ssim_abl}. It is interesting to note that Xception performs less effectively than ViT for the first three Mask SSIM range values; however, this changes when considering the last ranges, indicating the presence of high-quality deepfakes with more localized artifacts.  The same pattern is also observed for EFNB4~\cite{efn_net}+SBI~\cite{sbi} and other CNN methods, which further confirms our hypothesis (common CNNs such as~\cite{efn_net, xception} can capture subtle artifacts more effectively than plain ViT).

% \subsection{Robustness to Noise}
% We provide additional results when applying other perturbation methods to Table~\ref{tabl:ff_noise_auc} in the main paper. The results are reported in Table~\ref{tabl:ff_noise_auc_full}.

% \begin{table*}
% \caption{Robustness inspection on the FF++ to common perturbation methods.}
% \centering
% %\setlength\tabcolsep{2pt}
% \scalebox{0.85}{
% % \resizebox{\columnwidth}{!}{%
% \begin{tabular}{c|cccccc|c}
% \hline
% {Method} & Saturation & Contrast & Block & Noise & Blur & Pixel & Avg \\ 
% \hline
% \hline

% FakeFormer & 98.1 & 96.0 & 97.0 & 75.3 & 87.4 & 91.2 & 90.8 \\
% FakeSwin & 99.8 & 99.8 & 99.9 & 81.6 & 89.8 & 93.7 & 94.1 \\
% \hline
% \end{tabular}%
% }
% \label{tabl:ff_noise_auc_full}
% \end{table*}

% \subsection{Face-swaps versus Facial Reenactments: an Example}
% To illustrate the difference between the four manipulation methods used in FF++, we provide four examples, one for each method. It can be seen in Figure~\ref{fig:locality_levels} that the scale of artifacts in Face2Face and NeuralTextures is much subtler and intractable as compared to those in Deepfakes and FaceSwap. 

% \begin{figure}
%   \centering
%   \includegraphics[width=\linewidth]{figures/FF++_artifacts_old.pdf}
%   \caption{Example illustrating the four types of deepfakes in FF++~\cite{ff++}. It can be observed that Face2Face and NeuralTextures exhibit more subtle artifacts.}
%     \label{fig:locality_levels}
% \end{figure}

\subsection{Effect of Patch Size}

To further investigate the impact of the patch size on the performance, we provide, in addition to the experiment reported in Figure.~\textcolor{red}{3}-b of the main manuscript, the obtained cross-evaluation results of FakeFormer when varying the input size and the patch size in Table.~\ref{tabl:patch_size_abl}. It can be observed that either reducing the patch size or increasing the input resolution contributes to improving the generalization performance of the model. This confirms that the patch size and the input size implicitly affect the extraction of localized features that characterize deepfakes.

{
    \small
    \bibliographystyle{ieeenat_fullname}
    \bibliography{main}
}


\end{document}